\documentclass{article}

\usepackage[numbers]{natbib}

\usepackage[preprint]{neurips_2025}

\usepackage[utf8]{inputenc} 
\usepackage[T1]{fontenc}    
\usepackage{url}            
\usepackage{booktabs}       
\usepackage{amsfonts}       
\usepackage{nicefrac}       
\usepackage{microtype}      
\usepackage{xcolor}         
\usepackage[pdftex]{graphicx}
\usepackage{epstopdf}
\usepackage{multirow}
\usepackage{colortbl}
\usepackage[pagebackref=true,breaklinks=true,colorlinks,bookmarks=false,linkcolor=purple]{hyperref}
\usepackage{subcaption}
\usepackage[namelimits]{amsmath} 
\usepackage{amssymb}            
\usepackage{mathrsfs}            
\usepackage{bm}
\usepackage{wrapfig}
\usepackage{bbm}
\usepackage{cleveref}
\usepackage{algorithm}
\usepackage{algorithmic}
\usepackage{marvosym}
\usepackage{inconsolata}

\definecolor{c2}{HTML}{FBD9BD}
\definecolor{lp}{HTML}{f7f5fc}
\definecolor{c3}{HTML}{fe793d}
\definecolor{c4}{HTML}{eedeb0}
\definecolor{pp}{HTML}{BC7FCD}
\definecolor{bb}{HTML}{CDE8E5}
\definecolor{rouse}{rgb}{0.981,0.961,0.941}

\title{Integrating Extra Modality Helps Segmentor Find Camouflaged Objects Well}

\author{Chengyu Fang$^{1,}$\thanks{Equal Contribution, $\dagger$ Corresponding Author, \Letter~Email Address}
 \,,
 Chunming He$^{2,*}$\,,
        Yuelin Zhang$^3$ \,, 
	\textbf{Longxiang Tang}$^1$\,, \\
 \textbf{Chenyang Zhu}$^1$\,, 
 \textbf{Yuqi Shen}$^1$\,,
  \textbf{Chubin Chen}$^1$\,,
  \textbf{Guoxia Xu}$^4$\,,
	\textbf{Xiu Li$^{1,\dagger}$}\\
	$^1$Tsinghua University, ~$^2$Duke University, \\
        $^3$The Chinese University of Hong Kong,
        $^4$ Nanjing University of Posts and Telecommunications
 \\
 \Letter~\textit{chengyufang.thu@gmail.com}
 }

\begin{document}

\maketitle

\begin{abstract}
Camouflaged Object Segmentation (COS) remains challenging because camouflaged objects exhibit only subtle visual differences from their backgrounds and single-modality RGB methods provide limited cues, leading researchers to explore multimodal data to improve segmentation accuracy. In this work, we presenet MultiCOS, a novel framework that effectively leverages diverse data modalities to improve segmentation performance. MultiCOS comprises two modules: Bi-space Fusion Segmentor (BFSer), which employs a state space and a latent space fusion mechanism to integrate cross-modal features within a shared representation and employs a fusion‐feedback mechanism to refine context‐specific features, and Cross-modal Knowledge Learner (CKLer), which leverages external multimodal datasets to generate pseudo‐modal inputs and establish cross‐modal semantic associations, transferring knowledge to COS models when real multimodal pairs are missing. When real multimodal COS data are unavailable, CKLer yields additional segmentation gains using only non‐COS multimodal sources. Experiments on standard COS benchmarks show that BFSer outperforms existing multimodal baselines with both real and pseudo‐modal data. Code will be released at \href{https://github.com/cnyvfang/MultiCOS}{GitHub}.
\end{abstract}

\section{Introduction} \label{introduction}

Camouflaged Object Segmentation (COS) aims to detect hard-to-identify targets within a scene. The lack of clear visual cues and minimal contrast between camouflaged objects and their backgrounds make COS a challenging task for both machines and humans. To address the limitations of single-image, recent studies have explored the integration of cues from additional modalities to enhance performance. The use of multimodal data in COS can be broadly categorized into the following types.

\textcolor{pp}{\textbf{Type 1:}} Image-modality pairs collected using real-world sensors. For example, IPNet \cite{wang2024ipnet} and PolarNet \cite{wang2023polarization} utilize RGB-polarization image pairs captured from real-world camouflage scenes, enhancing segmentation by exploiting polarization cues. Nevertheless, these datasets remain limited in scale, and models trained on such sparse data often yield only marginal improvements in performance.

\textcolor{pp}{\textbf{Type 2:}} Image-modality pairs generated using established modality estimation techniques. Methods such as PopNet \cite{wu2023source}, DaCOD \cite{wang2023depth}, and DSAM \cite{yu2024exploring} use pretrained depth estimation models like NeWCRFs \cite{yuan2022neural} to infer depth maps from RGB camouflage images. These estimated pseudo-modality serve as auxiliary guidance to assist segment camouflaged objects. However, modality estimation is not always stable in some special scenario, and its results carry a certain degree of uncertainty. Insufficiently robust feature fusion methods are prone to be misled by erroneous estimation results.

\textcolor{pp}{\textbf{Type 3:}} In addition, some modalities are recognized for their potential in object-centered segmentation tasks. For example, infrared can capture thermal radiation differences, which help distinguish camouflaged objects. However, integrating these modalities into COS remains challenging. Constructing real-world paired datasets is difficult, and there are currently no reliable methods for generating pseudo-modality for camouflaged images. These limitations hinder the effective application of infrared and similar modalities in COS tasks. Therefore, we aim to design a framework that can fully exploit multimodal information under different data availability conditions to enhance COS tasks.

\begin{figure*}[t]
\setlength{\abovecaptionskip}{2px}
	\centering
	\includegraphics[width=\linewidth]{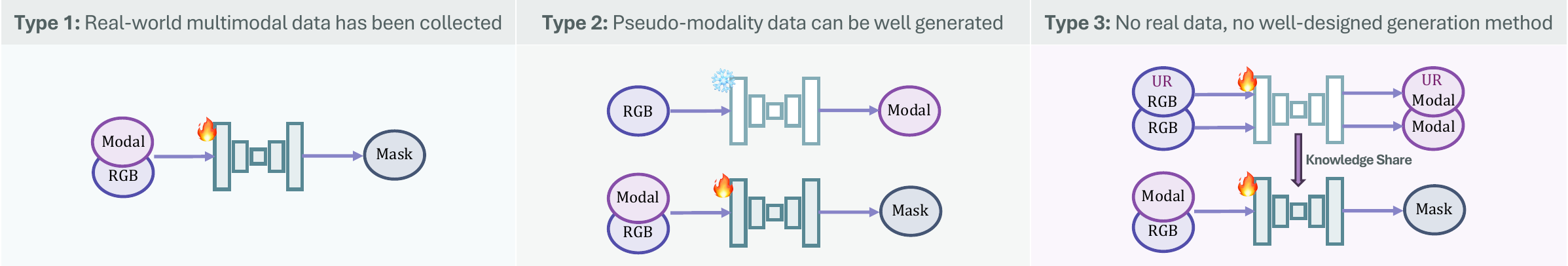}
	\caption{Training across different data scenarios: real data, generated data, and our MultiCOS. UR-RGB and UR-Modal means task-unrelated RGB image and corresponding multimodal data.
    }
	\label{fig:intro}
    \vspace{-5mm}
\end{figure*}

When paired RGB-modality data is available or high-quality pseudo-modal data can be reliably generated, the goal is to effectively integrate complementary cues from multiple modalities to enhance the model’s ability to distinguish camouflaged objects from complex backgrounds. To achieve this, we propose the Bi-space Fusion Segmentor (BFSer), which adopts a fusion-feedback-fusion strategy to progressively refine multimodal representations. Specifically, BFSer first uses a Latent Space Fusion Module (LSFM) for initial feature integration and a Feature Feedback Module (FFM) to guide modality-specific encoders with task-aware feedback. To further enhance global fusion, we introduce the State Space Fusion Mechanism (SSFM) and Cross State Space Model (CSSM), which align modality-specific features in a unified space, enabling coherent long-range dependency modeling and robust multimodal representation for improved MCOS performance.

When paired RGB-modality camouflaged data and reliable methods for generating high-quality pseudo-modal data are unavailable, we propose the Cross-modal Knowledge Learner (CKLer) to leverage multimodal datasets from other non-camouflaged tasks to learn cross-modal knowledge and enhance COS Tasks. CKLer generates both pseudo-modality and a semantic latent vector to guide the segmentation network, and is jointly optimized with it to enhance both segmentation and translation quality. Its modularity allows integration with existing segmentation networks, serving as a plug-and-play component to boost performance with the guidance of cross-modal knowledge.

\textbf{Our contributions can be summarized as follows:
}

(1) We introduce MultiCOS, a unified MCOS framework that combines a multimodal segmentor (BFSer) with a cross-modal knowledge learning module (CKLer). Extensive experiments demonstrate that our approach achieves state-of-the-art performance while offering plug-and-play versatility.

(2) We propose BFSer, which fuses features by latent and state space modules, and leverages a feedback mechanism to guide further extraction, enhancing contextual understanding and robustness.

(3) We propose CKLer, which acquires cross-modal knowledge from task-unrelated multimodal datasets by mapping images into the target modal space to generate pseudo-modal content and a knowledge vector, embedding this into the segmentor to establish cross-modal semantic associations.

\section{Related Works} \label{related_works}

\noindent \textbf{Camouflaged Object Segmentation}. Recent studies on COS have progressed using techniques such as multi-scale \cite{pang2024zoomnext}, multi-space \cite{zhong2022detecting, sun2025frequency}, multi-stage \cite{jia2022segment}, and biomimetic strategies \cite{he2023strategic}, which focus on enhancing information extraction from camouflaged images. Despite these advancements, most methods still rely on single-modal inputs, which limits the potential of multimodal data due to challenges in acquiring paired multimodal data with camouflaged samples. Advances in depth estimation have encouraged the integration of depth data, underscoring the benefits of multimodal approaches \cite{wu2023source, wang2023depth, yu2024exploring, xiang2022exploringdepthcontributioncamouflaged, wang2024depth}. However, research into RGB-to-X modal translation for other modalities is still limited, which restricts the advancement of additional modality-assisted COS tasks.

To address this issue, we propose CKLer to learns and utilizes cross-modal information between images and various modalities to enhance MCOS performance. By embedding a cross-modal semantic vector into the segmentor and leveraging existing non-camouflaged multimodal data
, this framework improves COS performance when real multimodal datasets with camouflaged objects are unavailable.

\noindent \textbf{State Space Models}. Rooted in classical control theory \cite{10.1115/1.3662552}, State Space Models (SSMs) are essential for analyzing continuous long-sequence data. The Structured State Space Sequence Model (S4) \cite{gu2022efficientlymodelinglongsequences} initially modeled long-range dependencies, recently, Mamba \cite{gu2024mambalineartimesequencemodeling, xiao2025mambatree} introduced a selection mechanism  that enables the model to extract relevant information from the inputs. Mamba has been applied effectively in image restoration \cite{guo2024mambairsimplebaselineimage, li2024fouriermambafourierlearningintegration, yang2024learning, zheng2024fd, zheng2024u}, segmentation \cite{wang2024mamba, xing2024segmamba}, and other domains \cite{zhang2024motion, zubic2024state}, achieving competitive results. 
In the context of image fusion, approaches like MambaDFuse \cite{li2024mambadfusemambabaseddualphasemodel} and FusionMamba \cite{xie2024fusionmambadynamicfeatureenhancement} have leveraged Mamba to improve performance. However, these methods utilize SSMs only for feature extraction, neglecting the cross-modal state space features and Mamba’s selection capabilities across different modal features in a unified space.
To address this, we propose a universal State Space Fusion Mechanism that integrates and selectively extracts features across modalities within a unified space, enhancing MCOS performance.

\section{Methodology} \label{methodology}

\subsection{Preliminaries}

State Space Models (SSMs) describe sequential dynamics using a hidden state governed by linear ordinary differential equations:
\begin{equation}
h'(t) = Ah(t) + Bx(t), \quad y(t) = Ch(t),
\end{equation}
which are discretized for deep learning. A common discretization method is zero-order hold, yielding
\begin{equation}
\label{eq:ZOH}
\overline{A} = \exp(\Delta A), \quad 
\overline{B} = (\Delta A)^{-1}(\exp(\Delta A) - I)\Delta B,
\end{equation}
and producing the discrete-time recurrence:
\begin{equation}
\label{eq:discret-ssm}
h_k = \overline{A}h_{k-1} + \overline{B}x_k, \quad y_k = Ch_k.
\end{equation}
To improve efficiency, Mamba \cite{gu2022efficientlymodelinglongsequences} approximates \(\overline{B} \approx \Delta B\) using a first-order Taylor expansion. Moreover, Mamba introduces input-dependent parameterization of \(B\), \(C\), and \(\Delta\), allowing the model to dynamically adapt to input content and capture complex long-range dependencies beyond the constraints of linear time-invariant systems.

\subsection{BFSer: Bi-space Fusion Segmentor}

\begin{figure*}[t]
\setlength{\abovecaptionskip}{0cm}
	\centering
	\includegraphics[width=\linewidth]{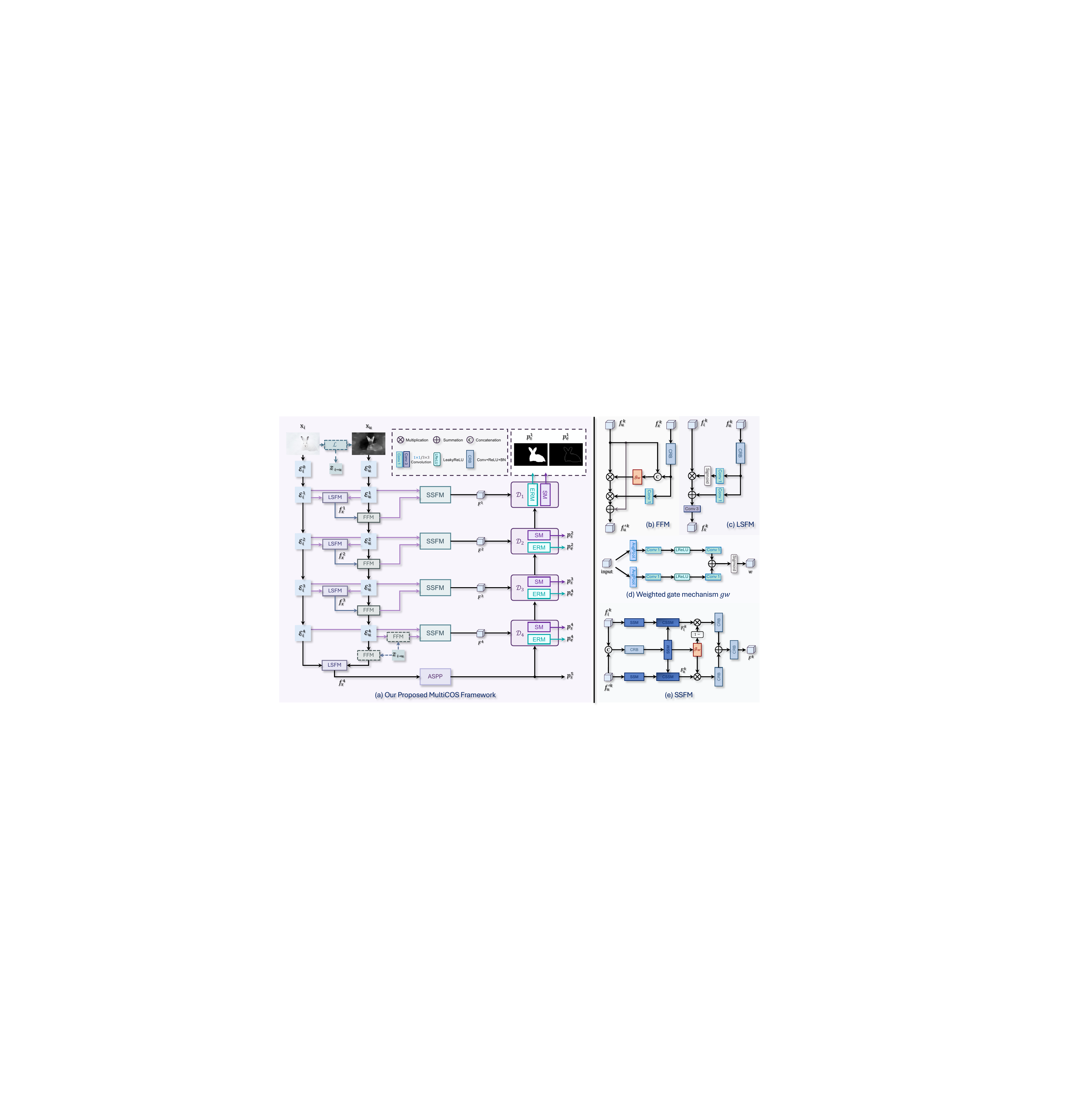}
	\caption{Framework of our MultiCOS, and the details of FFM, LSFM, \(g_w\), and SSFM. The modules outlined by dashed lines mean the modules introduced by CKLer.
    }
	\label{fig:Framework}
	\vspace{-5mm}
\end{figure*}

BFSer integrates features from RGB images and additional modalities within both the state space and the latent space.
The framework employs a Latent Space Fusion Module (LSFM) and a State Space Fusion Mechanism (SSFM) to selectively combine features from RGB images and auxiliary modalities, enhancing the performance of camouflaged object segmentation. 
Furthermore, a Feature Feedback Module (FFM) is introduced to leverage the outputs of LSFM at specific network layers, guiding subsequent encoder layers toward more effective feature extraction.

\subsubsection{MultiModal Segmentation-Oriented Encoder} 

BFSer conducts a two-branch encoder architecture to extract and utilize the features beneficial from different modalities. Give inputs \(\mathbf{x}_i\) and \(\mathbf{x}_u\), we first interpolate them to a uniform size of \(W \times H\). 
We begin by using a basic encoder \(\mathcal{E}_i\) to extract a set of deep features \(\{f_i^k\}_{k=0}^4\) from \(\mathbf{x}_i\), where each \(f_i^k\) has a resolution of \(\frac{W}{2^{k+1}} \times \frac{H}{2^{k+1}}\). 
To handle features from the additional modality, a secondary encoder \(\mathcal{E}_u\) with a similar architecture is employed. This encoder includes a customized embedding layer to adapt to the specific characteristics of \(\mathbf{x}_u\). 
The output of layer \(k\) of \(\mathcal{E}_u\) is denoted as \(f_u^k\), with the same resolution as \(f_i^k\).

To fuse features from different modalities in the latent space, we implement LSFM to fuse features \(f_i^k\) and \(f_u^k \in \mathbb{R}^{B \times C \times H \times W}\), generating a fused latent feature \(f_x^k\) of the same size at $k=\{1,2,3,4\}$:
\begin{equation}
\label{eq:LSFM}
f_x^k = f_i^k \odot \mathrm{Sigmoid}(W_c^1 \mathcal{C}(f_u^k)) + W_c^2 \mathcal{C}(f_u^k),
\end{equation}
where \(W_c\) is a convolution, $\mathcal{C}$ means a $\text{Conv+LReLU+BN}$ block, \(\odot\) means elementwise multiplication.

The last fused latent feature map \(f_x^4\), which is rich in semantic content, is processed by an atrous spatial pyramid pooling (ASPP) module \(A_s\) \cite{yang2018denseaspp} to produce a coarse prediction \(p_s^5=W_c(A_s(f_x^4)\), with the spatial resolution of \(f_x^4\) and serving as the initial point for the decoder.

Different from \(f_x^4\), the purpose of \(\{f_x^k\}_{k=1}^3\) is to guide $\mathcal{E}_u$ to extract targeted features from extra-modal by existing feature. 
To achieve this, BFSer introduces FFM to inject \(f_x^k\) into \(f_u^k\) in a gated way, generating \(f_u^{k'}\).
This updated feature serves as an input for both the $(k+1)^{th}$ layer of $\mathcal{E}_u$ and the SSFM following layer $k$:
\begin{equation}
        \alpha = \mathrm{Sigmoid}(W_c^1 \mathrm{conca}\Big[f_u^k,\, \mathcal{C}_1(f_x^k)\Big]), ~~
    f_u^{k'} = \mathcal{C}_2((f_u^k \odot \alpha \odot W_c^2\mathcal{C}_1(f_x^k)) + \mathrm{f_u^k})
\end{equation}
For a robust feature fusion, we propose SSFM, which selectively integrates features from different modalities within a unified state space representation:
\begin{equation}
\label{eq:SSFM_input}
    F^k = \mathrm{SSFM}(W_c^1f_i^k, W_c^2f_u^{k'}),
\end{equation}
where \(W_c^1f_i^k\), \(W_c^2f_u^{k'} \in \mathbb{R}^{B \times d_{m} \times H \times W} \), and \(\{F^k\}_{k=1}^4\) providing more complete context, reducing redundancy, filtering out noise, and capturing relationships between modalities. In the decoding stage, each layer of the decoder takes \(F^k\) as a conditional input. Combined with \(p_s^5\) reconstructed using \(A_s\) and features fused through the latent space, these inputs collectively enrich the reconstruction process by providing detailed and modality-aware information.

\subsubsection{Details of SSFM}
\textbf{State Space Fusion Mechanism}
In the vision state space model with a two-dimensional selective scan module, the feature is flattened into a sequence and scanned in four directions 
(top-left to bottom-right, bottom-right to top-left, top-right to bottom-left, and bottom-left to top-right) 
to capture the long-range dependencies of each sequence using the discrete state space equation. 
We propose the Cross State Space Model to facilitate information interaction between different sequences within the state space.

\begin{figure*}[t]
\setlength{\abovecaptionskip}{0cm}
	\centering
	\includegraphics[width=\linewidth]{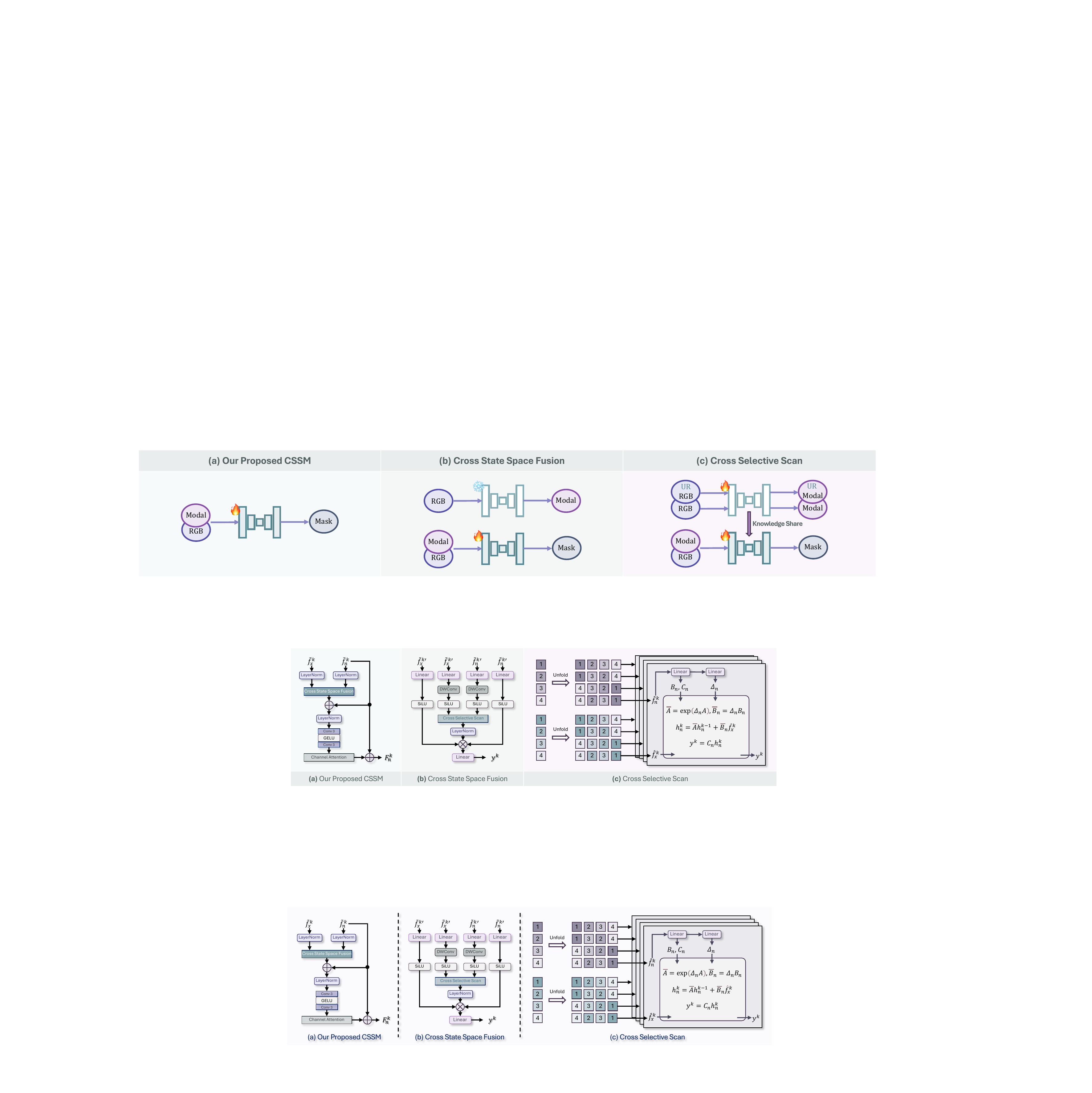}
	\caption{Details of our proposed CSSM.}
	\label{fig:CSSM}
	\vspace{-5mm}
\end{figure*}

After reshape \(f_i^k, f_u^{k'}\)
in \cref{eq:SSFM_input} to \(\mathbb{R}^{B \times H \times W \times d_{m}}\).We implement the vision state space module (SSM) as a residual state space block, as demonstrated by \cite{guo2024mambairsimplebaselineimage}, and utilize it as a form of long-range self-attention to process \(f_i^k\) and \(f_u^{k'}\), calculating the intra-modal correlation:
\begin{equation}
    \tilde{f}_i^k = SSM(f_i^k),\quad \tilde{f}_u^k = SSM(f_u^{k'}),\\
\end{equation}
then we process the self-modal correlation and cross-modal correlation with cross state space model (CSSM) we proposed to further fuse the bi-modals features in state space:
\begin{equation}
\label{eq:SSM-CSSM}
    \begin{aligned}
    \tilde{f}_x^k = SSM(\mathcal{C}(\mathrm{conca}(f_i^k, f_u^{k'}))), ~
    F_i^k = CSSM(\tilde{f}_i^k, \tilde{f}_x^k), ~ F_u^k = CSSM(\tilde{f}_u^k, \tilde{f}_x^k)
    \end{aligned}
\end{equation}
We utilize a weighted gate mechanism \(g_w\) to merge the transformed features as follows:
\begin{equation}
    \begin{aligned}
        F^k = \mathcal{C}(\mathcal{C}(g_wF_i^k + \tilde{f}_x^k) + &\mathcal{C}((1-g_w)F_u^{k} + \tilde{f}_x^k)), \\
        g_w = \mathrm{Sigmoid}(\lambda_g\mathrm{conca}\Big[\delta_1,\delta_2\Big] + \mu_g), & ~~
        \delta_1 = \mathcal{F}(\tilde{f}_x^k, \theta),\, \delta_2= \mathcal{F}(\tilde{f}_x^k + \delta_1, \theta).
    \end{aligned}
\end{equation}
This gate mechanism balances the contributions of \(F_i^k\) and \(F_u^{k}\) based on the guidance from \(\tilde{f}_x^k\). The function \(\delta_1=\mathcal{F}(\tilde{f}_x^k, \theta)\) and \(\delta_2=\mathcal{F}(\tilde{f}_x^k + \delta_1, \theta)\) generate intermediate signals that influence the final fused feature. The \(\mathrm{Sigmoid}\) ensures that \(g_w\) remains between 0 and 1, thus regulating the relative contributions of each path to the output \(F^k\).

\textbf{Cross State Space Model.} Let the input be \(\tilde{f}_n^k, \tilde{f}_x^k \in \mathbb{R}^{B \times H \times W \times d_m}\), where $\tilde{f}_n^k$ can be $\tilde{f}_i^k$ or $\tilde{f}_u^k$ in \cref{eq:SSM-CSSM}.
 We first apply a linear projection to extend the channel dimension of \(\tilde{f}_n^k\) and \(\tilde{f}_x^k\) to $d \times 2$ and split them along the last dimension into two parts:$\tilde{f}_n^{k'}, \tilde{f}_x^{k'} $,and $ z_n^k, z_x^k  \in \mathbb{R}^{B \times H \times W \times d}$. 

Next, we regard \(\tilde{f}_n^{k'}, \tilde{f}_x^{k'}\) as having the shape \(\mathbb{R}^{B \times d \times H \times W}\) and apply a depthwise convolution with a kernel size of \(d_{\text{conv}}\), followed by a nonlinear activation:
\begin{equation}
        \hat{f}_n^k = \mathrm{SiLU}\!\bigl(W_c^1(\tilde{f}_n^{k'})\bigr), \quad\hat{f}_x^k = \mathrm{SiLU}\!\bigl(W_c^2(\tilde{f}_x^{k'})\bigr).
\end{equation}
Here, the number of convolution groups equals the channel dimension \(d\), \(\mathrm{SiLU}\) is the activation function, and \(W_c\) means the convolutional layer. To fuse the two modalities in state space, we rewrite the \cref{eq:ZOH} and \cref{eq:discret-ssm} with:
\begin{equation}
    \overline{A} = \exp \bigl(\Delta_n A\bigr), \quad
    \overline{B}_n = \Delta_n B_n, \quad
    h^k_n = \overline{A}h^{k-1}_n \,+\, \overline{B}_n\hat{f}_x^k, \quad
    y^k = C_nh^k_n ,
\end{equation}
where the $B_n$, $C_n$, and $\Delta_n$ mean matrices $B$, $C$, and $\Delta$ with the selective mechanism parameters $sB(\hat{f}_n^k)=\mathrm{Linear_N}(\hat{f}_n^k)$, and $sC(\hat{f}_n^k)=\mathrm{Linear_N}(\hat{f}_n^k)$.

After combining the four directional sequences, we apply a layer normalization to \(y^k\) and then multiply it elementwisely by the activation of \(z_n^k\) and \(z_x^k\):
\begin{equation}
y'^k \;=\; \mathrm{LayerNorm}(y^k) \;\odot\; \mathrm{SiLU}(z_n^k) \;\odot\; \mathrm{SiLU}(z_x^k),
\end{equation}
we map \(y'^k\) back to the desired output dimension:
\begin{equation}
Y^k \;=\; y'^k \, W_{l} \;+\; b_{l},
\end{equation}
where \(W_{l} \in \mathbb{R}^{d \times d_{m}}\), \(b_{l} \in \mathbb{R}^{d_{m}}\), and \(Y^k \in \mathbb{R}^{B \times H \times W \times d_{m}}\).

Finally, to enhance the expressive capacity of different channels, we incorporate a Channel Attention mechanism (\(CA\)) within the CSSM to reduce channel redundancy. Additionally, we employ two weighted residual connections with \(s\) and \(s' \in \mathbb{R}^{C}\) to improve the network’s robustness:
\begin{equation}
    F^k = CA(W_c(\mathrm{LayerNorm}(Y^k+s\tilde{f}_n^k)))+ s'\tilde{f}_n^k
\end{equation}

\subsubsection{Replaceable Segmentation Decoder}

As our encoder employs a plug-and-play design, the decoder in BFSer can be substituted with any decoder that utilizes a coarse result or latent map and skip connections as inputs. 

In our implementation, we uses a multi-task segmentation decoder, such as ICEG \cite{he2023strategic}. 
This decoder features separate task heads for segmentation and edge reconstruction at each layer, with edge reconstruction providing additional supervision. 
The decoding process can be formulated as:
\begin{equation}
    \{p_s^k\}_{k=1}^4,\{p_e^k\}_{k=1}^4=\mathcal{D}(p_s^5, \{F^k\}_{k=1}^4),
\end{equation}
where \(\mathcal{D}\) means the decoder, \(\{p_s^k\}_{k=1}^4\) and \(\{p_e^k\}_{k=1}^4\) denote the results and reconstructed edges.

\subsubsection{Optimization}

As a unified plug-and-play method, our MultiModal Segmentation-Oriented Encoder with multi-space fusion can easily integrate with most non-specialized input design decoders.
Here, we use the multi-task segmentation decoder, which we employ as the default, as an example.

Our BFSer employs the weighted intersection-over-union loss \(L_{I}\), the weighted binary cross-entropy loss \(L_{B}\) to constrain the segmentation results \(\{p_s^k\}_{k=1}^5\), and the dice loss \(L_{D}\) to supervise the edge reconstruction results \(\{p_e^k\}_{k=1}^4\). Let the segmentation \(\mathbf{y_s}\) and edge \(\mathbf{y_e}\) as ground-truth, the total loss of BFSer can be presented as:
\begin{equation}
L_{\mathcal{S}}=\sum\nolimits_{k=1}^5\frac{1}{2^{k-1}}\left(L_{B}\left(p_s^k,\mathbf{y_s}\right)+L_{I}\left(p_s^k,\mathbf{y_s}\right)\right)+\sum\nolimits_{k=1}^4\frac{1}{2^{k-1}}L_{D}\left(p_e^k,\mathbf{y_e}\right).
\end{equation}

\subsection{CKLer: Cross-Modal Knowledge Learning}
CKLer \(\mathcal{L}\) is a plug-in encoder-decoder-like network. When the COS dataset lacks corresponding multimodal data, CKLer enables learning the mapping between images and modalities by introducing additional non-COS multimodal datasets, thereby aiding the COS task.

Specifically, we denote the images of the introduced additional dataset as \(\mathbf{e_i}\) and the corresponding additional modal data as \(\mathbf{e_u}\). We expect \(\mathcal{L}\) to learn the mapping relationship between them and obtain:
\begin{equation}
    \mathbf{\dot{e_u}} = \mathcal{L}(\mathbf{e_i}), \quad \mathbf{\dot{e_u}} \rightarrow \mathbf{e_u}.
\end{equation}
When working in collaboration with BFSer, CKLer inputs the image \(\mathbf{x_i}\) and, through the encoding and decoding process, obtains the corresponding pseudo-modality \(\mathbf{x_u}\) as well as the latent vector \(z_{\mathbf{i}\rightarrow\mathbf{u}}\) that embodies the knowledge of mapping between image and modality:
\begin{equation}
        \mathbf{x_u} = \mathcal{L}_\mathcal{D}(z_{\mathbf{i}\rightarrow\mathbf{u}})
,\quad z_{\mathbf{i}\rightarrow\mathbf{u}} = \mathcal{L}_\mathcal{E}(\mathbf{x_i}),
\end{equation}
where \(\mathcal{L}_\mathcal{E}\) and \(\mathcal{L}_\mathcal{D}\) are the encoder and decoder of \(\mathcal{L}\), \(z_{\mathbf{i}\rightarrow\mathbf{u}}\) means the latent vector which contains the knowledge of the map from \(\mathbf{x_i}\) to \(\mathbf{x_u}\). 

To integrate the \(z_{\mathbf{i}\rightarrow\mathbf{u}}\) to guide the segment process, we inject it to BFSer at \(k=4\) by replacing the LSFM(\cref{eq:LSFM}) with a new formula:
\begin{equation}
f_x^4 = f_i^4 \odot \mathrm{Sigmoid}(W_c^1 \mathcal{C}(FFM(f_u^4, z_{\mathbf{i}\rightarrow\mathbf{u}}))) + W_c^2 \mathcal{C}(FFM(f_u^4, z_{\mathbf{i}\rightarrow\mathbf{u}})),
\end{equation}
This operation integrates the mapping information between image and pseudo-modality, along with the semantic information extracted from both modalities, into the latent space. This unified representation strengthens the segmentation by leveraging complementary cross-modal knowledge.

\subsubsection{Optimization}
When employing CKLer, we perform joint training of CKLer and BFSer, optimizing the parameters of both networks using a shared optimizer. To enable CKLer to learn the mapping between \(\mathbf{e_i}\) and \(\mathbf{e_u}\), we utilize an L1 norm loss, formulated as:
\begin{equation}
    L_{\mathcal{L}} = ||\dot{e_u} - e_u||_1
\end{equation}
The total loss \(L_t\) for this joint training setup is expressed as: 
\begin{equation}
    L_t = L_{\mathcal{S}} + L_{\mathcal{L}}
\end{equation}

\begin{table*}[htp]
		\setlength{\abovecaptionskip}{0cm} 
		\setlength{\belowcaptionskip}{-0.2cm}
		\centering
            \caption{Quantitative comparisons of MultiCOS$\dagger$ (RGB-I) and other 12 SOTAs with two different type of backbones. {\color[HTML]{6f30a0}\textbf{Purple}}  and {\color[HTML]{578793}\textbf{cyan}} indicate the best and the second best.} \label{table:CODQuanti}
            \vspace{1mm}
            \setlength{\tabcolsep}{1.4mm}
		    \resizebox{\columnwidth}{!}{
			\begin{tabular}{l|cccc|cccc|cccc|cccc} 
				\toprule
				\multicolumn{1}{c|}{}      & \multicolumn{4}{c|}{\textit{CHAMELEON} }                                                                                                                                         & \multicolumn{4}{c|}{\textit{CAMO} }                                                                                                                                             & \multicolumn{4}{c|}{\textit{COD10K} }                                                                                                                                          & \multicolumn{4}{c}{\textit{NC4K} }                                                                                                                        \\ \cline{2-17} 
				\multicolumn{1}{l|}{\multirow{-2}{*}{Methods}} & {\cellcolor{gray!20}$M$~$\downarrow$}                                  & {\cellcolor{gray!20}$F_\beta$~$\uparrow$}                               & {\cellcolor{gray!20}$E_\phi$~$\uparrow$}                               & \multicolumn{1}{c|}{\cellcolor{gray!20}$S_\alpha$~$\uparrow$}                                   & {\cellcolor{gray!20}$M$~$\downarrow$}                                  & {\cellcolor{gray!20}$F_\beta$~$\uparrow$}                               & {\cellcolor{gray!20}$E_\phi$~$\uparrow$}                               & \multicolumn{1}{c|}{\cellcolor{gray!20}$S_\alpha$~$\uparrow$}                                   & {\cellcolor{gray!20}$M$~$\downarrow$}                                  & {\cellcolor{gray!20}$F_\beta$~$\uparrow$}                               & {\cellcolor{gray!20}$E_\phi$~$\uparrow$}                               & \multicolumn{1}{c|}{\cellcolor{gray!20}$S_\alpha$~$\uparrow$}                                   & {\cellcolor{gray!20}$M$~$\downarrow$}                                  & {\cellcolor{gray!20}$F_\beta$~$\uparrow$}                               & {\cellcolor{gray!20}$E_\phi$~$\uparrow$}                               & \multicolumn{1}{c}{\cellcolor{gray!20}$S_\alpha$~$\uparrow$}                                   \\ \midrule 
				\multicolumn{17}{c}{CNNs-Based Methods (ResNet50 Backbone)}                                   \\ \midrule

				\multicolumn{1}{l|}{SINet~\cite{fan2020camouflaged}}                   & 0.034                                 & 0.823                                 & 0.936                                 & \multicolumn{1}{c|}{0.872}                                 & 0.092                                 & 0.712                                 & 0.804                                 & \multicolumn{1}{c|}{0.745}                                 & 0.043                                 & 0.667                                 & 0.864                                 & \multicolumn{1}{c|}{0.776}                                 & 0.058                                 & 0.768                                 & 0.871                                 & 0.808                                 \\
				\multicolumn{1}{l|}{LSR~\cite{lv2021simultaneously}}                                          & 0.030                                 & 0.835                                 & 0.935                                 & \multicolumn{1}{c|}{0.890}                                 & 0.080                                 & 0.756                                 & 0.838                                 & \multicolumn{1}{c|}{0.787}                                 & 0.037                                 & 0.699                                 & 0.880                                 & \multicolumn{1}{c|}{0.804}                                 & 0.048 & 0.802 & 0.890                                 & 0.834                                 \\
				\multicolumn{1}{l|}{SLT-Net~\cite{cheng2022implicit}}                   & 0.030                                 & 0.835                                 & 0.940                                 & \multicolumn{1}{c|}{0.887}                                 & 0.082                                 & 0.763                                 & 0.848                                 & \multicolumn{1}{c|}{0.792}                                 & 0.036                                 & 0.681                                 & 0.875                                 & \multicolumn{1}{c|}{0.804}                                 & 0.049                                 & 0.787                                 & 0.886                                 & 0.830                                 \\
				\multicolumn{1}{l|}{SegMaR-1~\cite{jia2022segment}}                                    & \color[HTML]{578793}\textbf{0.028}& 0.828                                 &  0.944 & \multicolumn{1}{c|}{{0.892}} & 0.072 &0.772 & 0.861 & 0.805 & 0.035                                 & 0.699                                 & 0.890 & \multicolumn{1}{c|}{0.813}                                 & 0.052                                 & 0.767                                 & 0.885                                 & 0.835                                 \\
				\multicolumn{1}{l|}{OSFormer~\cite{pei2022osformer}}                                     & \color[HTML]{578793}\textbf{0.028} & 0.836 & 0.939                                 & \multicolumn{1}{c|}{0.891}                                 & 0.073                                 & 0.767                                 & 0.858                                 & \multicolumn{1}{c|}{0.799}                                 & 0.034 & 0.701 & 0.881                                 & \multicolumn{1}{c|}{0.811}                                 & 0.049                                 & 0.790                                 & 0.891 & 0.832                                 \\
	
                \multicolumn{1}{l|}{FEDER~\cite{He2023Camouflaged}}   & \color[HTML]{578793}\textbf{0.028} & \color[HTML]{578793}\textbf{0.850} & 0.944 & \multicolumn{1}{c|}{0.892} &  \color[HTML]{578793}\textbf{0.070} & {0.775} & 0.870 & \multicolumn{1}{c|}{0.802} & 0.032 & 0.715 & 0.892 & \multicolumn{1}{c|}{0.810} & {{0.046}} & {{0.808}} & {{0.900}} & {{0.842}} \\
                \multicolumn{1}{l|}{FGANet~\cite{zhaiexploring}}                                        & 0.030                                 & 0.838                                 & \color[HTML]{578793}\textbf{0.945}                                 & 0.891                                 & \color[HTML]{578793}\textbf{0.070}  & 0.769  & 0.865  & \multicolumn{1}{c|}{0.800}  & 0.032   & 0.708 & 0.894  & \multicolumn{1}{c|}{0.803}                                 & 0.047                                & 0.800                                 & 0.891                                 & 0.837                                 \\
                \multicolumn{1}{l|}{FocusDiff~\cite{zhao2025focusdiffuser}}  & \color[HTML]{578793}\textbf{0.028} & 0.843 & 0.938 & 0.890  & {\color[HTML]{6f30a0} \textbf{0.069}} & 0.772 & {\color[HTML]{6f30a0} \textbf{0.883}} &\color[HTML]{578793}\textbf{0.812} &\color[HTML]{578793}\textbf{0.031} &\color[HTML]{578793}\textbf{0.730} &\color[HTML]{578793}\textbf{0.897} & 0.820 &\color[HTML]{578793}\textbf{0.044} &\color[HTML]{578793}\textbf{0.810} &\color[HTML]{578793}\textbf{0.902} &\color[HTML]{578793}\textbf{0.850}    \\
                
                \multicolumn{1}{l|}{FSEL~\cite{sun2025frequency}}  & 0.029 & 0.847 & 0.941 &\color[HTML]{578793}\textbf{0.893}  & {\color[HTML]{6f30a0} \textbf{0.069}} & \color[HTML]{578793}\textbf{0.779} &\color[HTML]{578793}\textbf{0.881} & {\color[HTML]{6f30a0} \textbf{0.816}} & 0.032 & 0.722 & 0.891 &\color[HTML]{578793}\textbf{0.822} & 0.045 & 0.807 & 0.901 & 0.847    \\ 
                
                \rowcolor{lp} \multicolumn{1}{l|}{MultiCOS$\dagger$ }
                &{\color[HTML]{6f30a0} \textbf{0.024}} &{\color[HTML]{6f30a0} \textbf{0.866}} &{\color[HTML]{6f30a0} \textbf{0.951}} &{\color[HTML]{6f30a0} \textbf{0.902}} 
                
                &{\color[HTML]{6f30a0} \textbf{0.069}} &{\color[HTML]{6f30a0} \textbf{0.787}} &0.878 &{\color[HTML]{6f30a0} \textbf{0.816}} 
                
                &{\color[HTML]{6f30a0} \textbf{0.029}} &{\color[HTML]{6f30a0} \textbf{0.757}} &{\color[HTML]{6f30a0} \textbf{0.905}} &{\color[HTML]{6f30a0} \textbf{0.839}} 
                
                &{\color[HTML]{6f30a0} \textbf{0.042}} &{\color[HTML]{6f30a0} \textbf{0.820}} &{\color[HTML]{6f30a0} \textbf{0.910}} &{\color[HTML]{6f30a0} \textbf{0.857}} \\
                \midrule
                \multicolumn{17}{c}{Transformer-Based Methods (PVTv2 Backbone)}\\
                \midrule
				\multicolumn{1}{l|}{HitNet~\cite{hu2022high}}                                  & \color[HTML]{578793}\textbf{0.024}                                 & \color[HTML]{578793}\textbf{0.861}                                 & \color[HTML]{578793}\textbf{0.944}                                 & \multicolumn{1}{c|}{\color[HTML]{578793}\textbf{0.907}}                                 & 0.060                                 & 0.791                                 & 0.892                                 & \multicolumn{1}{c|}{0.834}                                 & 0.027                                 & 0.790                                 & 0.922                                 & \multicolumn{1}{c|}{0.847}                                 & 0.042                                 & 0.825                                 & 0.911                                 & 0.858                                 \\
                \multicolumn{1}{l|}{DaCOD~\cite{wang2023depth} }                &0.026 &0.829 &0.939 &0.893 &0.051 &0.831 &0.905 &\color[HTML]{578793}\textbf{0.855} &0.028 &0.740 &0.907 &0.840 &\color[HTML]{578793}\textbf{0.035} &0.833 &0.924 &0.874\\
                \multicolumn{1}{l|}{RISNet~\cite{wang2024depth} }                &--- &--- &--- &---  &\color[HTML]{578793}\textbf{0.050} &\color[HTML]{578793}\textbf{0.844} &\color[HTML]{578793}\textbf{0.922} &{\color[HTML]{6f30a0} \textbf{0.870}} &\color[HTML]{578793}\textbf{0.025} &\color[HTML]{578793}\textbf{0.804}  &\color[HTML]{578793}\textbf{0.931} &\color[HTML]{578793}\textbf{0.873} &0.037  &\color[HTML]{578793}\textbf{0.851} &\color[HTML]{578793}\textbf{0.925} &\color[HTML]{578793}\textbf{0.882} \\

                \rowcolor{lp} \multicolumn{1}{l|}{MultiCOS$\dagger$ } &{\color[HTML]{6f30a0} \textbf{0.019}} &{\color[HTML]{6f30a0} \textbf{0.884}} &{\color[HTML]{6f30a0} \textbf{0.962}} &{\color[HTML]{6f30a0} \textbf{0.920}} &{\color[HTML]{6f30a0} \textbf{0.048}} &{\color[HTML]{6f30a0} \textbf{0.845}} &{\color[HTML]{6f30a0} \textbf{0.923}} &{\color[HTML]{6f30a0} \textbf{0.870}} &{\color[HTML]{6f30a0} \textbf{0.021}} &{\color[HTML]{6f30a0} \textbf{0.809}} &{\color[HTML]{6f30a0} \textbf{0.933}} &{\color[HTML]{6f30a0} \textbf{0.874}} &{\color[HTML]{6f30a0} \textbf{0.032}} &{\color[HTML]{6f30a0} \textbf{0.859}} &{\color[HTML]{6f30a0} \textbf{0.932}} &{\color[HTML]{6f30a0} \textbf{0.887}} \\ 
    \bottomrule            
		\end{tabular}}
		\vspace{-3mm} 
\end{table*}

\begin{table*}[htp]
\centering
\setlength{\abovecaptionskip}{0cm}
\caption{Results on RGB-Depth COS. All the methods trained with source-free depth provided by \cite{wu2023source}}\label{Table:DCODQuanti}
\setlength{\tabcolsep}{1.4mm}
\resizebox{\columnwidth}{!}{ 
\begin{tabular}{l|cccc|cccc|cccc|cccc} 
            \toprule
            \multicolumn{1}{c|}{} &\multicolumn{4}{c|}{\textit{CHAMELEON} }   & \multicolumn{4}{c|}{\textit{CAMO} }      & \multicolumn{4}{c|}{\textit{COD10K} }     & \multicolumn{4}{c}{\textit{NC4K} }         \\ 
            \cline{2-17} 
            \multicolumn{1}{l|}{\multirow{-2}{*}{Methods}}  & {\cellcolor{gray!20}$M$~$\downarrow$}                                  & {\cellcolor{gray!20}$F^{x}_\beta$~$\uparrow$}                               & {\cellcolor{gray!20}$E^{x}_\phi$~$\uparrow$}                               & \multicolumn{1}{c|}{\cellcolor{gray!20}$S_\alpha$~$\uparrow$}                  & {\cellcolor{gray!20}$M$~$\downarrow$}                                  & {\cellcolor{gray!20}$F^{x}_\beta$~$\uparrow$}                               & {\cellcolor{gray!20}$E^{x}_\phi$~$\uparrow$}                               & \multicolumn{1}{c|}{\cellcolor{gray!20}$S_\alpha$~$\uparrow$}                                   & {\cellcolor{gray!20}$M$~$\downarrow$}                                  & {\cellcolor{gray!20}$F^{x}_\beta$~$\uparrow$}                               & {\cellcolor{gray!20}$E^{x}_\phi$~$\uparrow$}                               & \multicolumn{1}{c|}{\cellcolor{gray!20}$S_\alpha$~$\uparrow$}                                   & {\cellcolor{gray!20}$M$~$\downarrow$}                                  & {\cellcolor{gray!20}$F^{x}_\beta$~$\uparrow$}                               & {\cellcolor{gray!20}$E^{x}_\phi$~$\uparrow$}                               & \multicolumn{1}{c}{\cellcolor{gray!20}$S_\alpha$~$\uparrow$}                                   \\ 
            \midrule 
            CDINet \cite{zhang2021cross}  &0.036 &0.787 &0.903 &0.879    &0.100 &0.638 &0.766 &0.732    &0.044 &0.610 &0.821 &0.778     &0.067 &0.697 &0.830 &0.793 \\
            DCMF \cite{wang2022learning}  &0.059 &0.807 &0.853 &0.830      &0.115 &0.737 &0.757 &0.728    &0.063 &0.679 &0.776 &0.748     &0.077 &0.782 &0.820 &0.794 \\ 
            SPSN \cite{lee2022spsn}  &0.032 &0.866 &0.932 &0.887      &0.084 &0.782 &0.829 &0.773    &0.042 &0.727 &0.854 &0.789     &0.059 &0.803 &0.867 &0.813 \\ 
            DCF \cite{ji2021calibrated}   &0.037 &0.821 &0.923 &0.850    &0.089 &0.724 &0.834 &0.749    &0.040 &0.685 &0.864 &0.766     &0.061 &0.765 &0.878 &0.791 \\ 
            CMINet \cite{zhang2021rgb} &0.032 &0.881 &0.930 &0.891    &0.087 &0.798 &0.827 &0.782    &0.039 &0.768 &0.868 &0.811     &0.053 &0.832 &0.888 &0.839 \\ 
            SPNet \cite{zhou2021specificity} &0.033 &0.872 &0.930 &0.888 &0.083 &0.807 &0.831 &0.783    &0.037 &0.776 &0.869 &0.808     &0.054 &0.828 &0.874 &0.825 \\ 
            PopNet \cite{wu2023source} &\color[HTML]{578793}\textbf{0.022} &\color[HTML]{578793}\textbf{0.893} &\color[HTML]{578793}\textbf{0.962} &\color[HTML]{578793}\textbf{0.910}  &0.073 &0.821 &0.869 &0.806    &\color[HTML]{578793}\textbf{0.031} &0.789 &0.897 &0.827     &0.043 &0.852 &0.908 &0.852 \\ 
   
            DSAM \cite{yu2024exploring} & 0.028 & 0.877 & 0.957 & 0.883  &\color[HTML]{578793}\textbf{0.061} &\color[HTML]{578793}\textbf{0.834} &\color[HTML]{578793}\textbf{0.920} &\color[HTML]{578793}\textbf{0.832}    &0.033 &\color[HTML]{578793}\textbf{0.807} &\color[HTML]{578793}\textbf{0.931} &\color[HTML]{578793}\textbf{0.846}     &\color[HTML]{578793}\textbf{0.040} &\color[HTML]{578793}\textbf{0.862} &\color[HTML]{578793}\textbf{0.940} &\color[HTML]{578793}\textbf{0.871} \\              
            \rowcolor{lp}MultiCOS &{\color[HTML]{6f30a0} \textbf{0.018}} &{\color[HTML]{6f30a0} \textbf{0.912}} &{\color[HTML]{6f30a0} \textbf{0.970}} &{\color[HTML]{6f30a0} \textbf{0.923}}   
& {\color[HTML]{6f30a0} \textbf{0.048}} &{\color[HTML]{6f30a0} \textbf{0.865}} &{\color[HTML]{6f30a0} \textbf{0.929}} &{\color[HTML]{6f30a0} \textbf{0.867}}    
&{\color[HTML]{6f30a0} \textbf{0.020}} &{\color[HTML]{6f30a0} \textbf{0.850}} &{\color[HTML]{6f30a0} \textbf{0.946}} &{\color[HTML]{6f30a0} \textbf{0.880}}     
&{\color[HTML]{6f30a0} \textbf{0.031}} &{\color[HTML]{6f30a0} \textbf{0.882}} &{\color[HTML]{6f30a0} \textbf{0.944}} &{\color[HTML]{6f30a0} \textbf{0.890}}\\ 
            \bottomrule
    \end{tabular}}
    \vspace{-5mm}
\end{table*}

\section{Experiments} \label{experiments}
\subsection{Experimental Setup} \label{experimenal_setup1}
We evaluated our method on three multimodal COS tasks: RGB-Infrared (RGB-I), RGB-Depth (RGB-D), and RGB-Polarization (RGB-P). For RGB-I, we used datasets unrelated to COS to demonstrate CKLer’s ability to transfer knowledge from non-task-specific data. In the RGB-D task, pseudo-depth maps generated by well-trained depth estimation model were adopted, while the RGB-P task utilized real degree of linear polarization (DoLP) measurements. This setup enabled a comprehensive assessment of BFSer’s performance and robustness across both synthetic and real multimodal inputs.

CKLer uses a simple ResUNet backbone, which can be easily replaced.
BFSer employs PVTv2 \cite{wang2022pvt} as the default backbone, with additional results reported using ResNet50 \cite{he2016deep} for fair comparison. Detailed information about implementation, datasets, and evaluation metrics are provided in \ref{sec:impl}, \ref{sec:datasets}, and \ref{sec:metrics}. All results are computed using consistent task-specific evaluation protocols. MultiCOS 
\begin{wrapfigure}{r}{0.445\textwidth}
    \vspace{-2mm}
    \centering
    \setlength{\abovecaptionskip}{2pt}
    \setlength{\belowcaptionskip}{-1pt}
    {\footnotesize
    \captionof{table}{Quantitative comparisons of PCOD.}
    \label{table:PCODQuanti}
    \setlength{\tabcolsep}{2.5mm}
    \renewcommand{\arraystretch}{1}
    \resizebox{\linewidth}{!}{
        \begin{tabular}{l|cccc}\toprule 
        Methods   & \cellcolor{gray!20}$M \downarrow$ & \cellcolor{gray!20}$F_{\beta}^m \uparrow$ &\cellcolor{gray!20} $E_{\phi} \uparrow$& \cellcolor{gray!20}$S_{\alpha} \uparrow$\\
        \midrule
        SINet-V2 \cite{fan2021concealed}     & 0.013 & 0.819 & 0.941 & 0.882\\
        OCENet \cite{Liu_2022_WACV}       & 0.013 & 0.827  & 0.945 & 0.883\\
        ZoomNet \cite{pang2022zoom}      & 0.010 & 0.842 & 0.922 & 0.897\\
        BSANet \cite{zhu2022can}        & 0.011 & 0.861 & 0.945 & 0.903\\
        ERRNet \cite{ji2022fast}         & 0.023 & 0.704 & 0.901 & 0.833\\
        C2FNet-V2 \cite{chen2022camouflaged}  & 0.012 & 0.845 & 0.945 & 0.895\\
        PGSNet \cite{mei2022glass}        & 0.010 & 0.868 & 0.965 & 0.916\\
        CMX \cite{zhang2023cmx}         & 0.009 & 0.876 & 0.965 & 0.922\\
        DaCOD \cite{wang2023depth} &0.011 &0.846 &0.959 &0.899 \\
        IPNet \cite{wang2024ipnet}      & 0.008 & 0.882 & 0.970 & 0.922\\
        RISNet \cite{wang2024depth} &\color[HTML]{578793}\textbf{0.007} &\color[HTML]{578793}\textbf{0.904} &\color[HTML]{578793}\textbf{0.971} &\color[HTML]{578793}\textbf{0.933} \\
        \rowcolor{lp}MultiCOS      & {\color[HTML]{6f30a0} \textbf{0.006}}  & {\color[HTML]{6f30a0} \textbf{0.910}} & {\color[HTML]{6f30a0} \textbf{0.975}}  & {\color[HTML]{6f30a0} \textbf{0.937}} \\
		 \bottomrule  \end{tabular}
    }
    }

    \setlength{\abovecaptionskip}{0pt}
    \setlength{\belowcaptionskip}{0pt}
    {\footnotesize
    \captionof{table}{Quantitative results of RGB2IR.}
    \label{Table:i2iQuanti}
    \setlength{\tabcolsep}{2.5mm}
    \setlength{\tabcolsep}{0.8mm}
    \resizebox{\linewidth}{!}{
        \begin{tabular}{l|cccc}\toprule 
            Methods   & {\cellcolor{gray!20}PSNR~$\uparrow$} & {\cellcolor{gray!20}SSIM~$\uparrow$} & {\cellcolor{gray!20}LPIPS~$\downarrow$} & {\cellcolor{gray!20}RMSE~$\downarrow$} \\
            \midrule
            \multicolumn{5}{c}{M3FD-Detection Dataset} \\
            \midrule
            ResUNet & 0.742 & 19.82 & 0.491 & 27.56 \\
            \cellcolor{lp}CKLer (ResUNet) & \cellcolor{lp}\textbf{0.815} & \cellcolor{lp}\textbf{21.89} & \cellcolor{lp}\textbf{0.489} & \cellcolor{lp}\textbf{22.34} \\
        \bottomrule
        \end{tabular}
    }
    }
    \vspace{-1.2cm}
\end{wrapfigure}
refers to using BFSer only, while MultiCOS$\dagger$ means the combination of BFSer and CKLer.

\subsection{Quantitative and Qualitative Results} \label{compare}

\begin{figure*}[t]
\setlength{\abovecaptionskip}{2px}
	\centering
	\includegraphics[width=\linewidth]{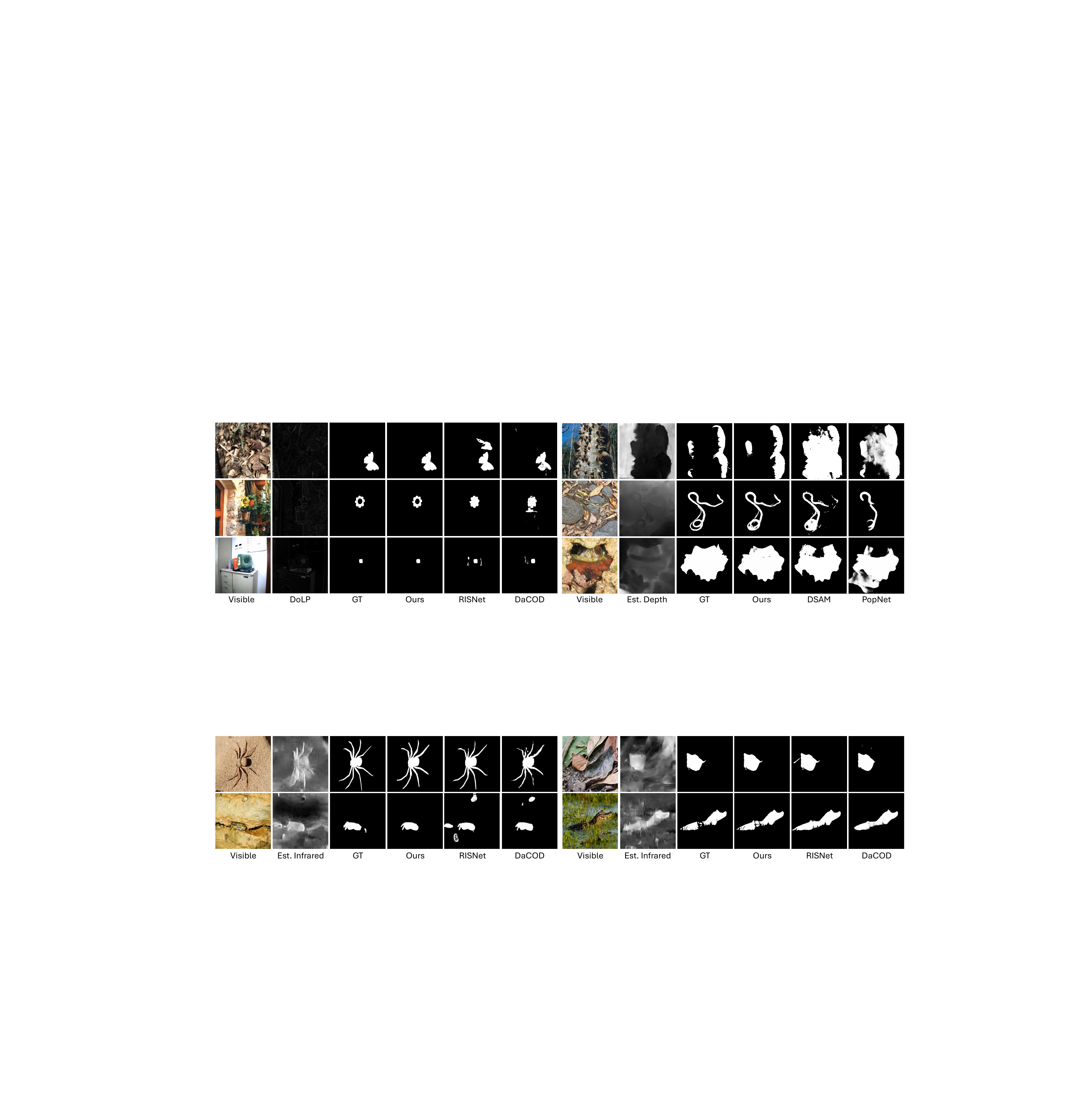}
	\caption{Qualitative results of MultiCOS$\dagger$ on RGB-I and other cutting-edge methods.}
	\label{fig:I-COMP}
    \vspace{-3mm}
\end{figure*}

\noindent \textbf{RGB and Task-Unrelated Infrared Data.} 
For the RGB-I task, MultiCOS$\dagger$ delivers strong performance by combining infrared and RGB inputs to improve camouflaged object segmentation. As shown in \cref{table:CODQuanti}, MultiCOS$\dagger$ consistently outperforms 12 state-of-the-art methods across four benchmarks. The model achieves the best scores on nearly all metrics. Infrared data enhances the model’s ability to distinguish subtle foreground-background differences, particularly in scenes with weak texture or lighting. Visual comparisons in \cref{fig:I-COMP} further confirm that the pseudo-infrared generated by our CKLer which trained jointly with BFSer using a task-unrelated RGB-I dataset, provides clear and precise segmentation boundaries, highlighting the advantage of our method under this scenario.

\begin{figure*}[t]
\setlength{\abovecaptionskip}{2px}
	\centering
	\includegraphics[width=\linewidth]{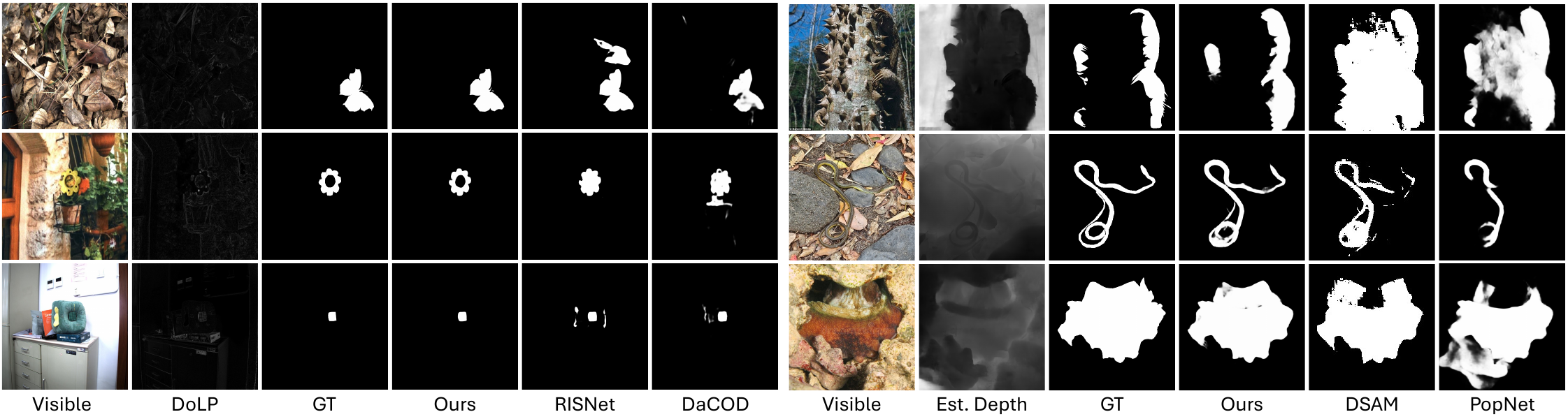}
	\caption{Qualitative results of MultiCOS on RGB-P and RGB-D and other cutting-edge methods.}
	\label{fig:D-P-COMP}
    \vspace{-5mm}
\end{figure*}

\textbf{Paired RGB and Pseudo-Depth Data.} 
In the RGB-D task, our MultiCOS model leverages pseudo-depth data paired with RGB images to effectively address the challenges of camouflaged object segmentation. 
Quantitative results presented in \cref{Table:DCODQuanti} 
demonstrate that MultiCOS outperforms 
competing methods, achieving the highest scores across all evaluated metrics. Additionally, visual comparisons in \cref{fig:D-P-COMP} highlight MultiCOS's capability to clearly distinguish foreground objects from their surroundings. Even in scenarios with minimal depth cues, as shown in the first row of \cref{fig:D-P-COMP}, MultiCOS still delivers superior segmentation performance and shows robustness. 

\textbf{Paired RGB and Real Polarization Data.}
For the RGB-P task, our MultiCOS model demonstrates exceptional performance by integrating real DoLP data with RGB imagery to improve the detection of camouflaged objects. 
As detailed in \cref{table:PCODQuanti}, MultiCOS achieves superior results on the PCOD1200 dataset. 
By leveraging polarization cues, the model uncovers details that are otherwise imperceptible to traditional RGB sensors. 
These cues are critical for precisely delineating object boundaries, as visually illustrated in \cref{fig:D-P-COMP}, where MultiCOS excels in segmenting subtle features and defining edges with remarkable precision. 
The success of MultiCOS in these complex scenarios highlights the significant advantages of incorporating polarization, enabling the detection of objects that would otherwise remain concealed in traditional imaging systems.

\subsection{Ablation Study} \label{ablation_study}

If not otherwise specified, we conduct ablation studies on \textit{COD10K} of the COS task.

\noindent \textbf{Effect of BFSer}. 
As illustrated in \cref{table:abl_bfser} and \cref{table:abl_bfser_sub}, BFSer significantly improves segmentation by integrating multimodal data. Removing either the extra modality encoder $\mathcal{E}_u$ or the image encoder $\mathcal{E}_i$ leads to a clear drop in accuracy, highlighting their essential roles. Likewise, excluding state-space fusion modules such as SSFM, CSSM, or LSFM negatively impacts performance, confirming their importance in ensuring robustness and accuracy. The removal of FFM also reduces performance, indicating its role in enhancing cross-stage feature 
integration. As shown in \cref{table:CompuCost}, these components yield notable gains with limited parameters.

\begin{wrapfigure}{r}{0.5\textwidth}
    \vspace{-8.5mm}
    \centering
    \setlength{\abovecaptionskip}{0pt}
    \setlength{\belowcaptionskip}{0pt}
    \includegraphics[width=\linewidth]{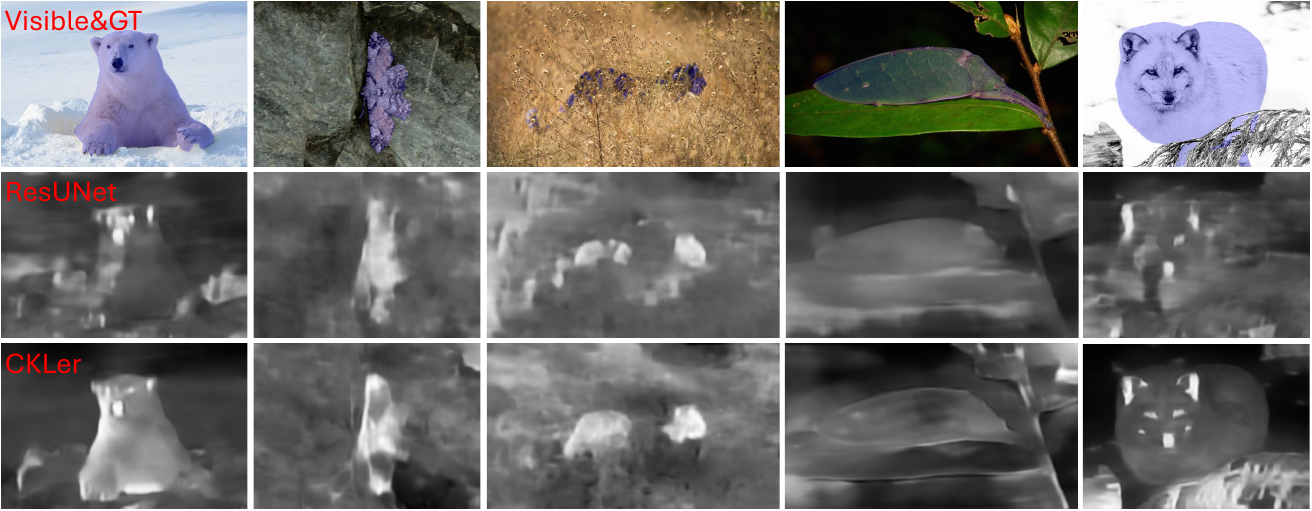}
    \caption{Qualitative results of RGB2IR.}
    \label{fig:IR}
    \vspace{-0.7cm}
\end{wrapfigure}

\noindent \textbf{Effect of CKLer}. Referencing \cref{table:abl_ckler}, CKLer enhances camouflaged object segmentation through the utilization of cross-modal knowledge. Disabling the 'Knowledge Injection' process, which involves integrating the latent vector $z_{i\rightarrow u}$, results in a noticeable decline in all metrics. This validates CKLer's efficacy in using extra multimodal data to improve the segmentation of camouflaged objects, enhancing both the accuracy and consistency of segmentation results across various datasets.

\begin{table}[t]
\centering
\setlength{\abovecaptionskip}{1px}
\begin{minipage}[t]{0.56\textwidth}
    \centering
    \setlength{\abovecaptionskip}{3pt}
    \setlength{\belowcaptionskip}{1pt}
    \captionof{table}{Effect of the BFSer.}
    \label{table:abl_bfser}
    \setlength{\tabcolsep}{1.9mm}
    \resizebox{\textwidth}{!}{
        \begin{tabular}{ccccc|cccc}\toprule 
        $\mathcal{E}_u$ &$\mathcal{E}_i$ &FFM &SSFM &LSFM & \cellcolor{gray!20}$M$~$\downarrow$ & \cellcolor{gray!20}$F^{x}_\beta$~$\uparrow$ & \cellcolor{gray!20}$E^{x}_\phi$~$\uparrow$ & \cellcolor{gray!20}$S_\alpha$~$\uparrow$\\
        \midrule
        $\times$ &$\checkmark$ &$\times$ &$\times$ &$\times$  &0.025 &0.770 &0.923 &0.867\\
        $\checkmark$ &$\checkmark$ &$\times$ &$\times$ &$\times$  &0.025 &0.783 &0.921 &0.868 \\
        $\checkmark$ &$\checkmark$ &$\times$ &$\checkmark$ &$\times$  &0.022 &0.796 &0.930 &0.876 \\
        $\checkmark$ &$\checkmark$ &$\times$ &$\times$ &$\checkmark$  &0.024 &0.789 &0.925 &0.871\\
        $\checkmark$ &$\checkmark$ &$\times$ &$\checkmark$ &$\checkmark$  &0.021 &0.812 &0.937 &0.880\\
        $\checkmark$ &$\checkmark$ &$\checkmark$ &$\times$ &$\checkmark$  &0.024 &0.792 &0.927 &0.873\\
        $\checkmark$ &$\checkmark$ &$\checkmark$ &$\checkmark$ &$\times$  &0.021 &0.802 &0.934 &0.877\\
        \rowcolor{lp}$\checkmark$ &\cellcolor{lp}$\checkmark$ &\cellcolor{lp}$\checkmark$ &\cellcolor{lp}$\checkmark$ &\cellcolor{lp}$\checkmark$  &\cellcolor{lp}\textbf{0.020} &\cellcolor{lp}\textbf{0.850} &\cellcolor{lp}\textbf{0.946} &\cellcolor{lp}\textbf{0.880}\\
        \bottomrule
        \end{tabular}
    }
\end{minipage}
\begin{minipage}[t]{0.433\textwidth}
    \centering
    \setlength{\abovecaptionskip}{3pt}
    \setlength{\belowcaptionskip}{1pt}
    \renewcommand{\arraystretch}{1.06}
    \captionof{table}{Results on more combinations.}
    \label{table:diff-combine}
    \setlength{\tabcolsep}{2mm}
    \resizebox{\textwidth}{!}{
        \begin{tabular}{l|cccc}\toprule 
        Methods   & \cellcolor{gray!20}$M \downarrow$ & \cellcolor{gray!20}$F_{\beta}^m \uparrow$ &\cellcolor{gray!20} $E_{\phi} \uparrow$& \cellcolor{gray!20}$S_{\alpha} \uparrow$\\
        \midrule
        \multicolumn{5}{c}{Validate on PCOD1200.}\\
        \midrule
        P Only &0.013 &0.826 &0.939 &0.884\\
        \cellcolor{lp}P-Est. D &\cellcolor{lp}\textbf{0.010} &\cellcolor{lp}\textbf{0.842} &\cellcolor{lp}\textbf{0.951} &\cellcolor{lp}\textbf{0.897}\\
        \midrule
        \multicolumn{5}{c}{Validate on COD10K.}\\
        \midrule
        RGB-Est. D &0.020 &0.850 &0.946 &0.880\\
        \cellcolor{lp}RGB-Est. P\&D &\cellcolor{lp}\textbf{0.019} &\cellcolor{lp}\textbf{0.861} &\cellcolor{lp}\textbf{0.952} &\cellcolor{lp}\textbf{0.887}\\
		 \bottomrule  \end{tabular}
    }
    
\end{minipage}
\vspace{-4mm}
\end{table}

\begin{table*}[t]
\centering
\setlength{\abovecaptionskip}{0.1cm}
\begin{minipage}[t]{0.493\textwidth}
    \centering
    \setlength{\abovecaptionskip}{3pt}
    \setlength{\belowcaptionskip}{1pt}
    \captionof{table}{Effect of the sub-modules.}
    \label{table:abl_bfser_sub}
    \resizebox{\textwidth}{!}{
        \begin{tabular}{ccc|cccc}\toprule 
        $g_w$ &SSM &CSSM & \cellcolor{gray!20}$M$~$\downarrow$ & \cellcolor{gray!20}$F^{x}_\beta$~$\uparrow$ & \cellcolor{gray!20}$E^{x}_\phi$~$\uparrow$ & \cellcolor{gray!20}$S_\alpha$~$\uparrow$\\
        \midrule
        $\times$ &$\checkmark$ &$\checkmark$ &0.021 &0.844 &0.943 &0.879\\
        $\checkmark$ &$\times$ &$\checkmark$ &0.021 &0.824 &0.936 &0.879\\
        $\checkmark$ &$\checkmark$ &$\times$ &0.023 &0.798 &0.931 &0.876\\ 
        \rowcolor{lp}$\checkmark$ &\cellcolor{lp}$\checkmark$ &\cellcolor{lp}$\checkmark$ &\cellcolor{lp}\textbf{0.020} &\cellcolor{lp}\textbf{0.850} &\cellcolor{lp}\textbf{0.946} &\cellcolor{lp}\textbf{0.880}\\
        \bottomrule
        \end{tabular}
    }

    \setlength{\abovecaptionskip}{3pt}
    \setlength{\belowcaptionskip}{1pt}
    \captionof{table}{Module computation cost.}
    \label{table:CompuCost}
    \renewcommand{\arraystretch}{0.91} 
    \resizebox{\textwidth}{!}{ 

        \begin{tabular}{l|cccc}
        \toprule 
        Modules &\cellcolor{gray!20}LSFMs &\cellcolor{gray!20}FFMs &\cellcolor{gray!20}SSFMs &\cellcolor{gray!20}ASPP \\
        \midrule
            \multicolumn{5}{c}{Run on RTX4090 with $448\times448$} \\
            \midrule
        \#Params. &1.161M &1.601M &7.539M &5.542M\\
        Time Cost &0.0011s &0.0008s &0.0193s &0.0012s\\
        \bottomrule  
        \end{tabular}
    }
\end{minipage}
\begin{minipage}[t]{0.499\textwidth}
    \centering
    \setlength{\abovecaptionskip}{3pt}
    \setlength{\belowcaptionskip}{1pt}
    \captionof{table}{Results on generalization of MultiCOS.}
    \label{table:generalization}
    \setlength{\tabcolsep}{3.5mm}
        \renewcommand{\arraystretch}{1.03} 
    \resizebox{\textwidth}{!}{
        \begin{tabular}{l|cccc} 
            \toprule
            Methods & {\cellcolor{gray!20}$M$~$\downarrow$} & {\cellcolor{gray!20}$F_\beta$~$\uparrow$} & {\cellcolor{gray!20}$E_\phi$~$\uparrow$} & {\cellcolor{gray!20}$S_\alpha$~$\uparrow$} \\ 
            \midrule 
            FEDER~\cite{He2023Camouflaged} & 0.032 & 0.715 & 0.892 & 0.810 \\
            \cellcolor{lp}FEDER$\dagger$ &\cellcolor{lp}\textbf{0.030} &\cellcolor{lp}\textbf{0.743} &\cellcolor{lp}\textbf{0.903} &\cellcolor{lp}\textbf{0.839} \\
            DaCOD~\cite{wang2023depth} & 0.028 & 0.740 & 0.907 & 0.840 \\
            \cellcolor{lp}DaCOD$\dagger$ &\cellcolor{lp}\textbf{0.025} &\cellcolor{lp}\textbf{0.783} &\cellcolor{lp}\textbf{0.929} &\cellcolor{lp}\textbf{0.856} \\
            \bottomrule
        \end{tabular}}

    \setlength{\abovecaptionskip}{3pt}
    \setlength{\belowcaptionskip}{1pt}
    \captionof{table}{Effect of the CKLer.}
    \label{table:abl_ckler}
    \setlength{\tabcolsep}{2.65mm}
    \resizebox{\textwidth}{!}{
        \begin{tabular}{l|cccc}
        \toprule 
        Methods & {\cellcolor{gray!20}$M$~$\downarrow$}                                  & {\cellcolor{gray!20}$F_\beta$~$\uparrow$}                               & {\cellcolor{gray!20}$E_\phi$~$\uparrow$}                               & \multicolumn{1}{c}{\cellcolor{gray!20}$S_\alpha$~$\uparrow$} \\
        \midrule
        w/o Know-Vec. &0.024 &0.792 &0.927 &0.869\\
        Only Know-Vec. &0.023 &0.795 &0.929 &0.873\\
        \cellcolor{lp}MultiCOS$\dagger$ &\cellcolor{lp}\textbf{0.021} &\cellcolor{lp}\textbf{0.809} &\cellcolor{lp}\textbf{0.933} &\cellcolor{lp}\textbf{0.874}\\
        \bottomrule  
        \end{tabular}}
\end{minipage}
\vspace{-3mm}
\end{table*}

\noindent \textbf{Effect of Downstream on Upstream.} As shown in \cref{fig:IR} and \cref{Table:i2iQuanti}, the results of RGB to infrared 
translation demonstrate that our joint training framework not only benefits the downstream task of camouflaged object segmentation but also improves upstream modality translation. 
Compared with ResUNet, the version enhanced by CKLer achieves better fidelity and perceptual score. This improvement confirms that downstream segmentation guidance benefits upstream feature learning and translation ability by enforcing structured supervision and improving cross-modal consistency. As shown in \cref{fig:IR}, our method produces sharper boundaries and more coherent infrared patterns, demonstrating the effectiveness of our training strategy and design in addressing the semantic challenges of RGB to Infrared translation.

\noindent \textbf{Generalization Analysis}. 
We evaluate the generalization ability of our proposed methods using the RGB-I task as a representative setting. As shown in \cref{table:generalization}, extending the native single-modal method FEDER \cite{He2023Camouflaged} into a multimodal version by incorporating our CKLer and BFSer fusion modules yields clear performance improvements. In addition, augmenting the existing multimodal method DaCOD \cite{wang2023depth} with our methods further enhances its results. These outcomes demonstrate the effectiveness of our module and training strategies in exploiting multimodal cues and highlight its robustness and flexibility as a plug-and-play framework for improving performance on segment camouflaged objects.

\noindent \textbf{Extension to More Combinations.} While most camouflaged object segmentation methods focus on RGB-based segmentation, our plug-and-play framework is designed to support non-RGB primary modalities. To demonstrate this, we modified and trained our model using only polarization and labels from the PCOD1200 dataset, along with depth-polarization pairs from the HAMMER dataset \cite{jung2022my}, which are not related to COS tasks. The results in \cref{table:diff-combine} show that combining polarization with depth leads to better segmentation performance on PCOD1200 than using polarization alone. Furthermore, our framework can be easily adapted to accommodate multiple modality combinations. By 
\begin{wrapfigure}{r}{0.445\textwidth}
    \vspace{-4mm}
    \centering
    \setlength{\abovecaptionskip}{2pt}
    \setlength{\belowcaptionskip}{-1pt}
    \setlength{\tabcolsep}{1.5mm}
    {\footnotesize
    \captionof{table}{Results on misaligned data.}
    \label{table:shift}
    \renewcommand{\arraystretch}{1}
    \resizebox{\linewidth}{!}{
        \begin{tabular}{l|cccc}\toprule 
        Methods   & {\cellcolor{gray!20}$M$~$\downarrow$}                                  & {\cellcolor{gray!20}$F_\beta$~$\uparrow$}                               & {\cellcolor{gray!20}$E_\phi$~$\uparrow$}                               & \multicolumn{1}{c}{\cellcolor{gray!20}$S_\alpha$~$\uparrow$}\\
        \midrule
        \multicolumn{1}{l|}{RGB Only} &0.025 &0.770 &0.923 &0.867 \\
        RGB\&Unaligned Infrared   &0.023 &0.798 &0.929 &0.871  \\
        RGB\&Unaligned Depth &0.021 &0.843 &0.944 &0.879 \\ 
		 \bottomrule
         \end{tabular}
    }
    }

    \setlength{\abovecaptionskip}{0pt}
    \setlength{\belowcaptionskip}{0pt}
    {\footnotesize
    \captionof{table}{Results of Simple baseline}\label{Table:Baseline}
    \setlength{\tabcolsep}{2.5mm}
    \resizebox{\linewidth}{!}{ 
    \begin{tabular}{l|cccc}\toprule 
            Methods   & {\cellcolor{gray!20}$M$~$\downarrow$}                                  & {\cellcolor{gray!20}$F_\beta$~$\uparrow$}                               & {\cellcolor{gray!20}$E_\phi$~$\uparrow$}                               & \multicolumn{1}{c}{\cellcolor{gray!20}$S_\alpha$~$\uparrow$}\\
            \midrule
            \multicolumn{1}{l|}{PVTv2 Baseline} &0.030 &0.732 &0.906 &0.854\\
            \cellcolor{lp}PVTv2 Baseline$\dagger$ &\cellcolor{lp}\textbf{0.026} &\cellcolor{lp}\textbf{0.762} &\cellcolor{lp}\textbf{0.918} &\cellcolor{lp}\textbf{0.863} \\ 
    		 \bottomrule  \end{tabular}}
        }
    \vspace{-0.8cm}
\end{wrapfigure}
introducing both depth and polarization estimation during training, 
we observe further improvements on COD10K, validating the framework’s flexibility and effectiveness across diverse input configurations.

\noindent \textbf{Robustness to Misaligned Data.} We simulate pixel misalignment by cropping the top-left 90\% of the auxiliary modality data and resizing it to the original size, introducing spatial shifts and inconsistency. As shown in \cref{table:shift}, even under such misalignment, incorporating unaligned inputs still outperforms the RGB-only baseline. This highlights the robustness of our method, which benefits from CKLer’s ability to capture cross-modal correspondences and BFSer’s feature-level fusion strategy.

\noindent \textbf{Improvement on Simple Backbone.}
To further verify the general applicability of our approach, we apply our modules to a simple baseline built on the PVTv2 backbone \cite{wang2022pvt}. As shown in \cref{Table:Baseline}, the enhanced baseline (denoted as PVTv2 Baseline$\dagger$) achieves consistent performance gains across all evaluation metrics compared to its original counterpart. This demonstrates that our method is effective even without strong architectural priors and can serve as a general enhancement strategy for various COS backbones.

\section{Limitations and Future Works} \label{limitation}
While MultiCOS has achieved outstanding results in RGB-X COS tasks, two limitations remain. 

\textit{1) The Bias Between CKLer and Modal Translation:} 
CKLer is designed to capture associative knowledge and mapping relationships between RGB and additional modalities, primarily to guide the BFSer segmentation network. 
Its focus is not on generating highly precise pseudo-modal information, which may result in outputs that deviate from traditional modality translation expectations. 
Further research is needed to improve the interpretability of these generative mechanisms and understand their contribution to segmentation performance.

\textit{2) Restricted Segmentation in Dual-Modal Scenarios:} 
At present, the application of SSMs and BFSer in MCOS is confined to dual-modality setups employing a dual-encoder architecture. 
However, leveraging the robust capabilities of SSMs in capturing long-range contextual dependencies, the framework holds promise for extension to support additional modalities, such as triple modalities or beyond, which could significantly enhance segmentation performance.

To further enhance segmentation performance, future efforts could focus on jointly fine-tuning existing pre-trained pre-processing models \cite{fang2024real,he2023reti,zhang2024unified}, translation networks \cite{fang2023joint}, refinement models \cite{ahn2021refining}, even the generative model \cite{zhu2024multibooth,zhu2024instantswap,wang2024taming,
  wang2024cove,he2024diffusion} alongside segmentation models \cite{xiao2024survey}, aiming to simultaneously improve the performance of both components. 
Additionally, leveraging multitask guidance to enhance RGB-X image translation, particularly for tasks that are challenging for conventional image-to-image translation methods, which emerges as a promising avenue for future research.

\section{Conclusions} \label{conclusions}
This work introduces MultiCOS for MCOS task. 
MultiCOS comprises BFSer, a multimodal segmentor, and CKLer, a cross-modal knowledge learning plugin, which cooperatively enhances segmentation accuracy. 
BFSer utilizes an SSFM and an LSFM to integrate cross-modal features each layer, along with an FFM to guide the encoding of subsequent layers, improving contextual understanding and reducing susceptibility to noise. Simultaneously, CKLer leverages multimodal data unrelated 
to the COS task to enhance model camouflaged objects segmentation capabilities by generating pseudo-modal content and learning cross-modal semantic knowledge. 
Our qualitative and quantitative results demonstrate that our MultiCOS outperforms existing MCOS approaches.

\newpage
\bibliographystyle{unsrt}
\bibliography{egbib}


\newpage
\appendix

\begin{figure*}[t]
\setlength{\abovecaptionskip}{0cm}
	\centering
	\includegraphics[width=\linewidth]{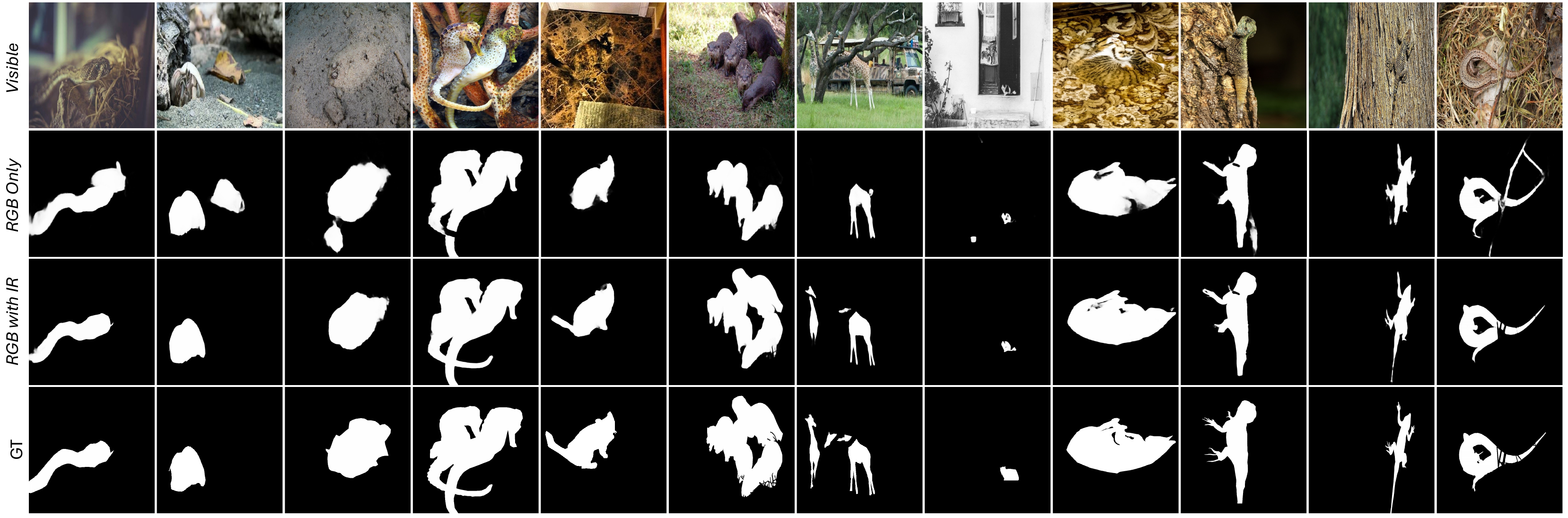}
	\caption{Qualitative results of RGB only and our RGB-Infrared results obtained through MultiCOS$\dagger$.}
	\label{fig:more-result}
\end{figure*}

\section{Methodology}

\subsection{More Details About SSMs}
\textbf{Structured State Space Sequence Models (S4)}. S4 transforms a one-dimensional input \(x(t) \in \mathbb{R}\) into an output \(y(t) \in \mathbb{R}\) through an implicit state representation \(h(t)\in \mathbb{R}^{N}\). The system dynamics are governed by the following linear ordinary differential equation:
\begin{equation}
\label{eq:ssm-sup}
    h'(t) = Ah(t) + Bx(t),\quad
    y(t)  = Ch(t),
\end{equation}
where \(N\) denotes the dimensionality of the hidden state. The matrices \(A \in \mathbb{R}^{N\times N}\), \(B \in \mathbb{R}^{N \times 1}\), and \(C \in \mathbb{R}^{1\times N}\) define the dynamics of the system and control how the hidden state evolves and how the output is derived.

To integrate \cref{eq:ssm-sup} into deep learning pipelines, the continuous formulation is typically discretized. Let \(\Delta\) denote a timescale step size that discretizes \(A\) and \(B\) into discretized \(\overline{A}\) and \(\overline{B}\). A common discretization approach is the zero-order hold, defined as:
\begin{equation}
\label{eq:ZOH-sup}
    \overline{A} = \exp (\Delta A),\,
    \overline{B} = (\Delta A)^{-1} (\exp(\Delta A) - I) \Delta B.
\end{equation}
By discretizing \cref{eq:ssm-sup} with the timestep 
\(\Delta\), the system is transformed into the following RNN-like representation: 
\begin{equation}
\label{eq:discret-ssm-sup}
    h_k = \overline{A}h^{k-1} + \overline{B}x^k,\quad
    y_k = Ch^k.
\end{equation}
where $h_k$ and $y_k$ represent the discretized hidden state and output, respectively, at timestep $k$.

In Mamba \cite{gu2022efficientlymodelinglongsequences}, the matrix \(\overline{B}\) can be approximated using the first-order Taylor series as follows: 
\begin{equation}
    \overline{B}\approx(\Delta A)(\Delta A)^{-1}\Delta B=\Delta B
\end{equation}

\noindent\textbf{Selective Scan Mechanism.} State Space Models (SSMs) are effective for modeling discrete sequences but are inherently constrained by their Linear Time-Invariant (LTI) nature, resulting in static parameters that remain unchanged regardless of input variations. The Selective State Space Model (S6, also known as Mamba) addresses this limitation by introducing input-dependent dynamics. In the design of Mamba, the matrices \( B \in \mathbb{R}^{L \times N} \), \( C \in \mathbb{R}^{L \times N} \), and \( \Delta \in \mathbb{R}^{L \times D} \) are directly derived from the input data \( x \in \mathbb{R}^{L \times D} \). This dependency allows the model to adapt dynamically to the input context, enabling it to capture complex interactions within long sequences more effectively. 

\section{Experiments} \label{sup_experiments}

\subsection{More Visualization Results} \label{sup_visualization}

We take RGB-IR as an example, As shown in \cref{fig:more-result}, infrared captures thermal cues, while RGB provides texture and color. Mapping both into the state space allows their features to interact, enhancing targets in low-contrast areas and adding fine details. Compared to ``RGB Only'', the ``RGB with IR'' setup which uses our SSFM and CKLer joint training improves object localization and contour sharpness, reducing errors in complex scenes.

\begin{figure*}[ht]
\setlength{\abovecaptionskip}{0cm}
	\centering
	\includegraphics[width=\linewidth]{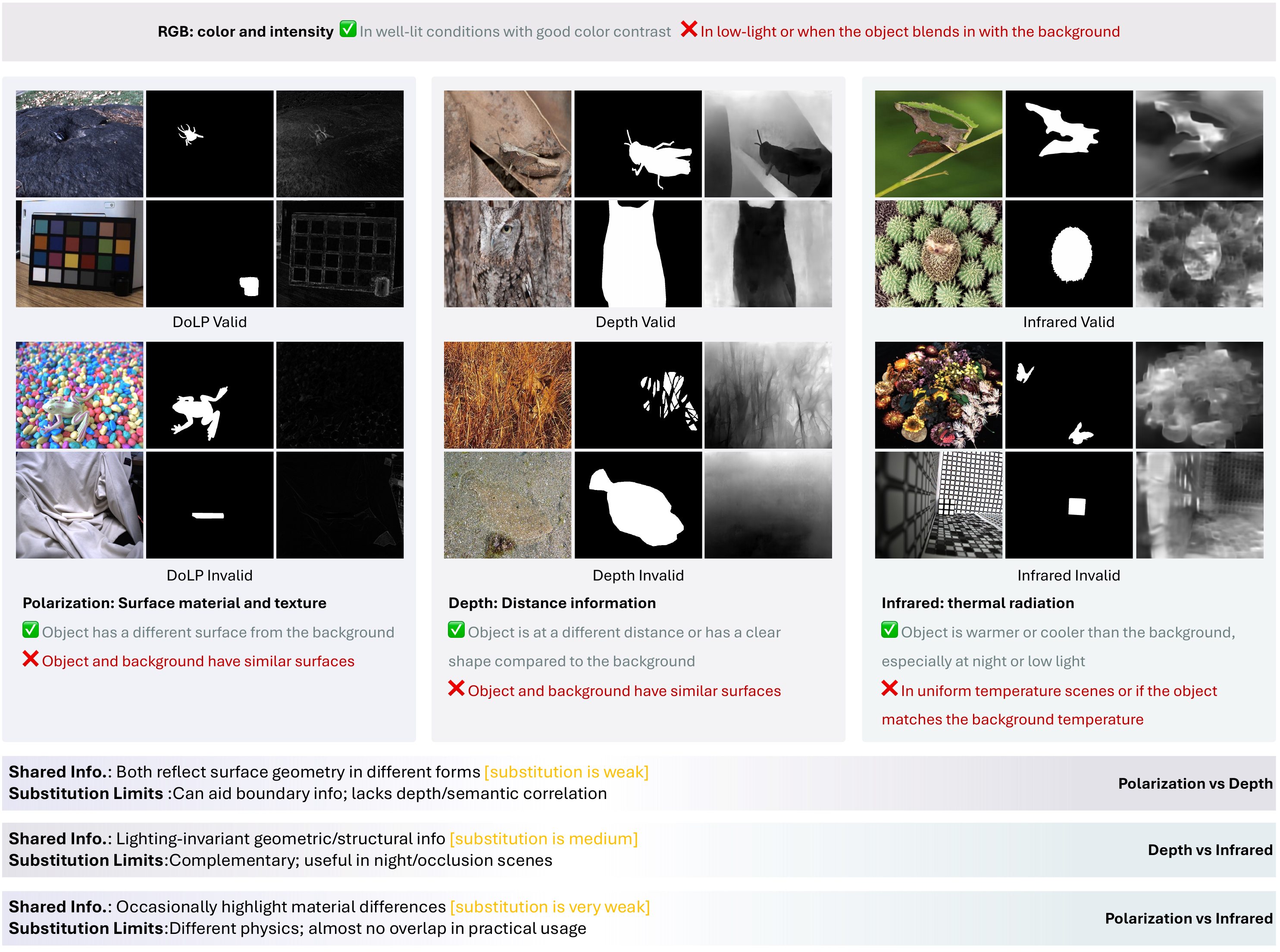}
	\caption{Examples of effective and ineffective cases in making the object easier to detect using DoLP, depth, and infrared modalities. From left to right: RGB image, ground truth, and corresponding modality visualization.}
	\label{fig:valid_invalid}
\end{figure*}

\subsection{Experimental Settings} \label{sup_settings}

\subsubsection{Implementation Details.} \label{sec:impl}  We implement our method in PyTorch and train our model on four RTX 4090 GPUs. We use the Adam optimizer with a learning rate of $1e-4$ and a batch size of 16. The input image size is $448 \times 448$, following \cite{wang2023depth}. We train the model for 160 epochs, with the learning rate gradually decaying to 5e-6. $d_m$ is set as 96, $d$ is set as 192, and \(d_{\text{conv}}\) is set as 3.

\subsubsection{Datasets} 
\label{sec:datasets}
Except for the RGB-P task, we employ the CHAMELEON \cite{skurowski2018animal}, CAMO \cite{le2019anabranch}, COD10K \cite{fan2021concealed}, and NC4K \cite{lv2021simultaneously} datasets for our evaluation. We follow the common setting of previous work, combining 3,040 pairs from COD10K with 1,000 pairs from CAMO to the training set. 

In the RGB-D task, to evaluate the performance of our methods under paired RGB with pseudo-modal data. we adopt the pseudo-depth map used in PopNet \cite{wu2023source} and DSAM \cite{yu2024exploring}, which paired with above four dataset, to fair comparison.

In the RGB-I task, to evaluate our MultiCOS in the scenario where an extra modality is missing, unlike the RGB-D task that uses a pseudo-depth map, we utilize the M3FD-Fusion dataset \cite{liu2022target} which contains 300 RGB-IR image pairs to allows our CKLer to learn and leverage cross-modal knowledge from the task unrelated RGB-Infrared data.

For the RGB-P task, we use the PCOD1200 dataset \cite{wang2024ipnet} to evaluate methods in the scenario with real multimodal data. This dataset contains 1,200 manually annotated pairs of RGB and DoLP (Degree of Linear Polarization) images. It is divided into 970 pairs for training and 230 pairs for testing.

\subsubsection{Metrics} 
\label{sec:metrics}
We use the different metrics on different tasks to fairly compare with previous works with the tasks common settings. 
The metrics we used include Mean Absolute Error (M), max F-measure ($F^x_\beta$), mean F-measure ($F^m_\beta$), adaptive F-measure ($F_\beta$), mean E-measure ($E_\phi$), max E-measure ($E^x_\phi$) and Structure Similarity ($S_\alpha$).

\section{Analysis the role of additional modality}

The additional modalities we incorporate, such as depth, infrared, and polarization, provide complementary information to visible images. Each captures distinct aspects of the scene: depth encodes geometric structure, infrared highlights thermal variations that are less influenced by camouflage, and polarization reveals surface properties. While all modalities contribute to scene understanding, their underlying physical signals differ, leading to variations in noise characteristics and the types of boundary cues they provide.

In \cref{tab:modality_valid} and \cref{fig:valid_invalid} we provide a comprehensive analysis, that details which modality types are effective under various conditions, why they work, and the common properties they share. In \cref{tab:modality_pairs} we specifically describe the differences and relationships between these modalities.
This analysis deepens the understanding of how complementary modalities improve COS task.

To handle these modality differences, we designed LSFM, FFM, and SSFM modules to effectively extract and fuse their complementary strengths. LSFM initially integrates RGB and auxiliary modality features in latent space, emphasizing useful signals. FFM feeds fused features back to the auxiliary modality encoder, guiding it to capture targets more accurately. SSFM integrates multimodal information in state space, capturing long-range dependencies and reducing noise. These lightweight modules require few parameters yet fully utilize complementary multimodal information, significantly improving camouflaged object segmentation performance.

\begin{table}[t]
\centering
\caption{Conditions under which each modality provides valid/invalid measurement.}
\label{tab:modality_valid}
\resizebox{\linewidth}{!}{
\begin{tabular}{l|l}
\toprule
\textbf{Measurement} & \cellcolor{gray!20}\textbf{Valid Condition} \\
\midrule
RGB & Color and intensity in well-lit conditions with good color contrast. \\ \midrule
Polarization & Surface material and texture. objects have a different surface from the background. \\ \midrule
Infrared & Thermal radiation. objects are warmer or cooler than the background, especially at night or in low light. \\ \midrule
Depth & Distance information. objects are at a different distance or have a clear shape compared to the background. \\
\midrule
\textbf{Measurement} & \cellcolor{gray!20}\textbf{Invalid Condition} \\
\midrule
RGB & In low-light or when the object blends in with the background. \\ \midrule
Polarization & Objects and background have similar surfaces. \\ \midrule
Infrared & In uniform temperature scenes or when the object matches the background temperature. \\ \midrule
Depth & Object and background are at similar distances or the scene has little depth variation. \\
\bottomrule
\end{tabular}
}
\vspace{-2mm}
\end{table}

\begin{table}[t]
\centering
\caption{Comparison of modality pairs in COS with shared information and substitution constraints.}
\label{tab:modality_pairs}
\setlength{\tabcolsep}{3.2mm}
\resizebox{\linewidth}{!}{
\begin{tabular}{l|l|l}
\toprule
\textbf{Modality Pair} & \cellcolor{gray!20}\textbf{Shared Information in COS} & \cellcolor{gray!20}\textbf{Notes on Substitution Limits} \\
\midrule
RGB vs IR & Edge/texture alignment in some cases & IR helps in low light, but lacks color and fine texture \\ \midrule
RGB vs Depth & Object shape and boundary cues & Depth aids segmentation in low-texture areas. lacks semantics \\ \midrule
RGB vs Polar. & Reflectance contrast, especially on shiny surfaces & Limited to special materials (e.g., metal, glass) \\ \midrule
IR vs Depth & Lighting-invariant geometric/structural info & Complementary. useful in night/occlusion scenes \\ \midrule
IR vs Polar. & Occasionally highlight material changes & Different physics, almost no overlap in practical usage \\ \midrule
Depth vs Polar. & Surface orientation cues & Can aid boundary info. lacks depth/semantic correlation \\
\bottomrule
\end{tabular}}
\end{table}

\section{Broader Impacts} \label{impacts}

Camouflaged Object Segmentation (COS) presents a significant challenge in computer vision due to the minimal visual distinction between foreground and background. Effective COS benefits a range of downstream applications such as search and rescue, autonomous navigation, surveillance, and wildlife monitoring. Our proposed framework, MultiCOS, combines a Bi-space Fusion Segmentor (BFSer) and a Cross-modal Knowledge Learner (CKLer) to enhance segmentation performance by integrating multimodal cues, even when real paired multimodal COS data are unavailable. The ability of CKLer to learn from non-task-specific datasets expands the applicability of COS to scenarios where data collection is difficult. Furthermore, the plug-and-play design of our modules facilitates integration into existing systems. To date, COS technologies have not been linked to negative societal outcomes. Similarly, our MultiCOS framework, designed for technical and research use, does not introduce any foreseeable risks in terms of ethical or societal impact.

\end{document}